\documentclass[preprints,article,accept,pdftex,moreauthors]{Definitions/mdpi} 
\firstpage{1} 
\pubvolume{1}
\issuenum{1}
\articlenumber{0}
\pubyear{2025}
\copyrightyear{2025}
\datereceived{ } 
\daterevised{ } 
\dateaccepted{ } 
\datepublished{ } 
\hreflink{https://doi.org/} 

\Title{IceBench: A Benchmark for Deep Learning based Sea Ice Type Classification}

\TitleCitation{IceBench: A Benchmark for Deep Learning based Sea Ice Type Classification}

\Author{Samira Alkaee Taleghan $^{1,*}$, Andrew P. Barrett $^{2}$, Walter N. Meier $^{2}$, Farnoush Banaei-Kashani $^{1,*}$}

\AuthorNames{Samira Alkaee Taleghan, Andrew P. Barrett, Walter N. Meier, Farnoush Banaei-Kashani}

\address{%
$^{1}$ College of Engineering, Design and Computing, University of Colorado Denver, Denver, CO, USA; 
samira.alkaeetaleghan@ucdenver.edu, farnoush.banaei-kashani@ucdenver.edu\\
$^{2}$ National Snow and Ice Data Center, CIRES, University of Colorado Boulder, Boulder, CO, USA; 
andrew.barrett@colorado.edu, walt@colorado.edu
}

\isAPAStyle{%
       \AuthorCitation{Alkaee Taleghan, S., Barrett, A. P., Meier, W. N., Karimzadeh, M., \& Banaei-Kashani, F.}
}{%
\isChicagoStyle{%
        \AuthorCitation{Alkaee Taleghan, Samira, Andrew P. Barrett, Walter N. Meier, Morteza Karimzadeh, and Farnoush Banaei-Kashani.}
}{%
        \AuthorCitation{Alkaee Taleghan, S.; Barrett, A. P.; Meier, W. N.; Karimzadeh, M.; Banaei-Kashani, F.}
}
}

\corres{Correspondence: samira.alkaeetaleghan@ucdenver.edu (S.A.T.); farnoush.banaei-kashani@ucdenver.edu (F.B.K.)}

\abstract{Sea ice plays a critical role in the global climate system and maritime operations, making timely and accurate classification essential. However, traditional manual methods are time-consuming, costly, and have inherent biases. Automating sea ice type classification addresses these challenges by enabling faster, more consistent, and scalable analysis. While both traditional and deep learning approaches have been explored, deep learning models offer a promising direction for improving efficiency and consistency in sea ice classification. However, the absence of a standardized benchmark and comparative study prevents a clear consensus on the best-performing models. To bridge this gap, we introduce \textit{IceBench}, a comprehensive benchmarking framework for sea ice type classification. Our key contributions are threefold: First, we establish the IceBench benchmarking framework which leverages the existing AI4Arctic Sea Ice Challenge dataset as a standardized dataset, incorporates a comprehensive set of evaluation metrics, and includes representative models from the entire spectrum of sea ice type classification methods categorized in two distinct groups, namely, pixel-based classification methods and patch-based classification methods. IceBench is open-source and allows for convenient integration and evaluation of other sea ice type classification methods; hence, facilitating comparative evaluation of new methods and improving reproducibility in the field. Second, we conduct an in-depth comparative study on representative models to assess their strengths and limitations, providing insights for both practitioners and researchers. Third, we leverage IceBench for systematic experiments addressing key research questions on model transferability across seasons (time) and locations (space), data downscaling, and preprocessing strategies. By identifying the best-performing models under different conditions, IceBench serves as a valuable reference for future research and a robust benchmarking framework for the field. 
}


\begin{document}




\section{Introduction}

Sea ice plays a pivotal role in the global climate system, it significantly impacts maritime operations, affecting shipping routes, resource exploration, and coastal communities in polar regions. As such, timely and accurate classification of sea ice types is essential for a wide range of applications, from climate modeling to maritime safety \cite{Vihma}\cite{Bobylev}\cite{Sandven}. Traditionally, sea ice type classification has relied on manual ice charting methods \cite{Zakhvatkina}\cite{Dedrick}. Although ice charting has proven valuable, its reliance on human annotation makes the process time-consuming and costly. These limitations reduce its reliability, especially as the demand for large-scale, up-to-date, and accurate sea ice classification grows in response to accelerating climate change.

To address these challenges, automating sea ice classification has become increasingly important. Early methods relied on traditional machine learning, while deep learning captures intricate spatial and textural patterns directly from raw data. This shift has revolutionized sea ice type classification, enhancing accuracy, scalability, and real-time mapping capabilities \cite{Li2024}\cite{Huang2024}. Deep learning models require large amounts of data to train effectively and learn meaningful patterns. However, progress in deep learning-based sea ice type classification was initially hindered by the lack of publicly available datasets. Without sufficient labeled data, models struggled to generalize and reach their full potential. Recent efforts to develop and share large-scale datasets have addressed this limitation, providing valuable resources for training and evaluation \cite{Malmgren-Hansen}. As a result, deep learning models can better learn intricate features directly from raw data, reducing the need for manual feature engineering and further enhancing classification accuracy. 

While the availability of large-scale datasets has accelerated progress in deep learning-based sea ice classification, a major challenge remains: the lack of a standardized model benchmark framework. Various deep learning approaches have been developed using different datasets, preprocessing techniques, and model architectures. Although this diversity has fostered innovation, it has also made it difficult to systematically evaluate and compare model performance. Without a common benchmark, determining the most effective approaches and understanding their strengths and limitations across different conditions remain challenging. 

To address this issue, we introduce \textit{IceBench}, a comprehensive benchmarking framework for automated sea ice type classification. IceBench encompasses a diverse set of deep learning methods, categorized into pixel-based and patch-based classification approaches, each offering distinct advantages. By establishing a standardized evaluation framework, IceBench facilitates objective model comparisons, enhances reproducibility, and provides valuable insights into the most effective strategies for sea ice classification. Building on this standardized framework, IceBench offers several key benefits. First, it ensures that models are evaluated on a common ground, allowing researchers to assess performance consistently and practitioners to select the most suitable methods for different scenarios \cite{Gut2022}\cite{Dwivedi}. Second, it helps track advancements in deep learning techniques for sea ice type classification, offering insights into emerging trends and improvements. Third, IceBench facilitates the identification of state-of-the-art methods, serving as a reliable reference for future research and development. Lastly, it promotes reproducibility by providing clear guidelines on datasets, evaluation metrics, and experimental setup. 

This paper makes several key contributions to the field of sea ice type classification:
\begin{enumerate}
    \item We introduce a comprehensive benchmark that includes the existing AI4Arctic Sea Ice Challenge dataset as a standardized dataset\cite{ChallengeDataset}, evaluation metrics, and representative models for each classification category. This benchmark establishes a common ground for evaluation of current and future sea ice classification methods.
\item We conduct a detailed comparative study on existing sea ice classification models using IceBench. This study helps identify the strengths and weaknesses of different approaches, guiding future research directions.
\item We use IceBench to perform extensive experimentation toward addressing longlasting research questions in this field. Specifically, our investigation focuses on transferability of models across seasons (time) and locations (space), data downscaling alternatives, and data preparation strategies. We also perform a parameter sensitivity analysis to evaluate impact of various data parameters on model performance, including patch size and dataset size.
\end{enumerate}

In summary, this study provides insight into the factors that influence classification accuracy and the trade-offs, offering valuable guidance and tools for future research and practice in sea ice classification. IceBench is released as an open-source software that allows for convenient integration and evaluation of other sea ice type classification methods to facilitate reproducibility\footnote{The IceBench code is available at \url{https://github.com/UCD-BDLab/IceBench}}. 

The remainder of this paper is organized as follows. Section 2 reviews related work in sea ice type classification, focusing on both pixel-based and patch-based approaches and the application of deep learning techniques in this domain. Section 3 details the components of our benchmarking framework, including the datasets, evaluation metrics, and methods. In Section 4, we describe the experimental evaluation, covering the methodology, results, and behavioral analysis of the models under various conditions. Finally, Section 5 presents the discussion and conclusion, summarizing key findings and suggesting future research directions. 

\section{Related Work}
In this section, we review the related work in sea ice classification, under the two categories of patch-based classification and pixel-based classification (or segmentation). We also review relevant benchmarking efforts in related fields, as there is currently no established benchmarking framework specifically for sea ice type classification.

\subsection{Sea Ice Classification}
This subsection covers existing deep learning models developed for sea ice type classification, including patch-based and pixel-based approaches. We discuss various deep learning models that have been utilized to address the challenges unique to sea ice type, highlighting their strengths and limitations.

\subsubsection{Patch-based Sea Ice Type Classification}
In this section, we focus on patch-based approaches, where single label is assigned to an entire image region or patch, providing a single classification for the entire area rather than individual pixels. The development of deep learning in image classification has progressively shaped advancements in sea ice classification. Early convolutional neural networks (CNNs) laid the groundwork, demonstrating the effectiveness of hierarchical feature extraction from complex remote sensing data. Li et al. \cite{Li2017} employed a CNN to analyze Synthetic Aperture Radar (SAR) images from the Chinese Gaofen-3 satellite, enhancing classification accuracy by training the network with synthetic patches assembled from smaller ones, which effectively differentiated between sea ice and non-ice areas. Boulze et al. \cite{Boulze} developed a CNN model to classify different types of sea ice in Sentinel-1 SAR data, demonstrating the model's superior performance over traditional algorithms, thus showcasing CNN's computational efficiency.

Advanced integrations of CNN architectures have also shown remarkable success in enhancing sea ice classification. Han et al. \cite{Han} introduced a method using a 3D convolutional neural network (3D-CNN) integrated with a gray-level co-occurrence matrix (GLCM) for analyzing hyperspectral sea ice images. This method enhances classification by combining deep spectral-spatial features. Another innovative approach by Han et al. \cite{Han2020} combined 3D-CNN and Squeeze-and-Excitation (SE) networks with an SVM classifier to enhance the classification by effectively integrating spatial and spectral data features, optimizing feature weighting, and improving accuracy in scenarios with limited sample sizes.

Building on the foundational concept of CNNs, AlexNet \cite{Krizhevsky} transformed computer vision by showcasing the effectiveness of deep networks in image classification. It introduced Rectified Linear Units (ReLUs) for faster training and dropout to reduce overfitting, paving the way for more advanced architectures. Building on the foundational work of AlexNet, Xu et al. \cite{Xu2017} explored application of CNN-based transfer learning for classifying sea ice and open water from SAR imagery. Their study leverages pre-trained AlexNet models, which, when fine-tuned on SAR data, significantly enhance classification accuracy. 

VGG16 \cite{Simonyan} further advanced CNN design by increasing network depth and standardizing small 3×3 convolutional filters, leading to improved hierarchical feature extraction. Its uniform architecture allowed for deeper networks capable of capturing more complex representations. Unlike earlier CNNs, its increased depth enables the capture of more complex and abstract features through successive layers. Khaleghian et al. \cite{Khaleghian2021} evaluated various CNN architectures for sea ice classification and found that a modified VGG16 model, trained from scratch on an augmented dataset, outperformed others. 

As networks grew deeper, the challenge of vanishing gradients emerged. ResNet (Residual Network) \cite{He} addressed this issue by introducing skip connections, which enabled efficient training of significantly deeper networks while maintaining gradient flow. Song et al. \cite{Song2018} developed the Sea Ice Residual Convolutional Network (SI-ResNet), designed for classifying sea ice types using Sentinel-1 SAR imagery. This model enhances classification performance by integrating residual learning and employs an ensemble learning strategy to improve accuracy. Similarly, Lyu et al. \cite{Lyu2022} utilized the Normalizer-Free ResNet (NFNet) for classifying sea ice types from RADARSAT Constellation Mission (RCM) data. The NFNet configuration outperformed traditional Random Forest classifiers, proving effectiveness of this model in handling the complex characteristics of dual-polarized SAR data. Zhang et al. \cite{Zhang2021} introduced MSI-ResNet (Mini Sea Ice), a mini sea ice residual convolutional network that utilizes fully polarimetric SAR data from Gaofen-3 satellite. This model enhances classification performance by optimizing the use of different polarization combinations, proving its superiority over traditional SVM classifiers in discriminating sea ice types with high accuracy and kappa coefficients. Additionally, Chen et al. \cite{Chen2023} presented AM-ResNet, an attention-based multi-label classification network. By integrating a Squeeze-and-Excitation (SE) module to enhance feature representation and applying neural network pruning to reduce computational load, AM-ResNet improves the accuracy and efficiency of sea ice classification. Jiang et al. \cite{Jiang2022} developed a sea ice-water classification algorithm using RADARSAT-2 imagery with ResNet and regional pooling. This method integrates Iterative Region Growing with Semantics (IRGS) \cite{Yu2008} segmentation and ResNet features, achieving 99.67\% accuracy and producing detailed sea ice maps.

Further refining these ideas, DenseNet (Densely Connected Convolutional Network) \cite{Huang} introduced dense connectivity, maximizing feature reuse and improving gradient propagation. Its efficient parameter utilization and strengthened feature learning further optimized CNN performance, influencing applications in sea ice classification. Kruk et al. \cite{Kruk} developed a model using DenseNet121 to classify sea ice types from SAR imagery. Their model achieved high classification accuracies and demonstrated the model’s effectiveness in distinguishing different ice stages from RADARSAT-2 data. Additionally, Han et al. \cite{Han2022} introduced a novel sea ice classification method using a dual-branch DenseNet architecture integrated with a squeeze-and-excitation attention mechanism. This approach significantly enhanced classification accuracy by effectively leveraging the complementary characteristics of SAR and optical data, outperforming single-source and other data fusion methods. Nagi et al. \cite{Nagi} utilized DenseNet to automatically detect Marginal Ice Zones (MIZs) in RADARSAT-2 satellite images, showcasing the model's capability as a fixed feature extractor.

Overall, these diverse adaptations and enhancements of CNN based methods demonstrate the potential of advanced architectures in addressing the unique challenges of sea ice type classification. By leveraging different network designs and strategies, researchers have achieved significant improvements.

\subsubsection{Pixel-based Sea Ice Type Classification}
This section explores pixel-based approaches, where each pixel is classified individually, providing a fine-grained understanding of sea ice types. Unlike patch-based methods, which classify entire image regions, pixel-based classification enables precise scene analysis through segmentation. Segmentation is particularly valuable for generating high-resolution sea ice maps, supporting detailed monitoring of sea ice extent and changes over time.

The U-Net model \cite{Ronneberger}, originally developed for biomedical image segmentation, has been widely adopted for sea ice classification. Its encoder-decoder structure and skip connections effectively capture multi-scale features, enabling precise pixel-based classification from remote sensing imagery. Building on U-Net’s success, recent advancements in deep learning have focused on enhancing U-Net-based models, leading to notable improvements in accuracy and efficiency. Ren et al. \cite{Ren2020} and Huang et al. \cite{Huang2021} have both employed U-Net-based models aimed at classifying sea ice and open water in SAR images. \cite{Ren2020} focused on distinguishing between sea ice and open water, achieving precision and recall rates over 91\%, while \cite{Huang2021} developed their model to differentiate various sea ice types, such as multi-year and first-year ice, thereby enhancing classification accuracy through pixel-based classification. Further innovations include the introduction of multitask and attention mechanisms to improve segmentation quality. Cantu \cite{Cantu} developed a hierarchical multitask U-Net model specifically for automated sea ice mapping, which enhances consistency by simultaneously predicting sea ice concentration, types, and floe size. Similarly, Ren et al. \cite{Ren2021} implemented position and channel attention mechanisms within the Dual-Attention U-Net (DAU-Net) to refine the classification of sea ice and open water. 
Ji et al. \cite{Ji2022} enhanced the U-Net architecture by incorporating batch normalization layers and an adaptive moment estimation optimizer, significantly improving the segmentation of Arctic sea ice. Wang \cite{Wang} employed an enhanced U-Net architecture with integrated stacking models, combining multiple specialized U-Net classifiers to achieve highly accurate sea ice segmentation.

Another notable model in pixel-based is DeepLab \cite{Chen2017}, which extends beyond traditional segmentation architectures by leveraging atrous convolution. This technique enhances multi-scale feature extraction, improving its ability to handle complex segmentation tasks with greater precision. Pires de Lima et al. \cite{Lima2023} developed a sea ice segmentation algorithm for Sentinel-1 images using ResNets and Atrous Spatial Pyramid Pooling (ASPP), achieving high accuracy for ice and water segmentation and outperforming a baseline U-Net model, and adding characterization of uncertainty in model outputs \cite{deLimaUncertainty}. Sun et al. \cite{Sun2023} introduces CA-DeepLabV3+, an innovative model for sea ice extraction that enhances the original DeepLabv3+ by integrating a coordinate attention mechanism. This modification significantly improves feature representation across both channel and spatial dimensions, enabling the model to capture fine details and large-scale information about sea ice more effectively. Balasooriya et al. \cite{Balasooriya} discuss the implementation and performance of the DeepLabv3 model for in-situ segmentation of sea ice, highlighting its enhanced efficiency and suitability for real-time applications on mobile and embedded hardware platforms. Zhang et al. \cite{Zhang2022} introduce Ice-Deeplab, a specialized adaptation of DeepLabv3+ for sea ice segmentation. This model incorporates a Convolution Block Attention Module (CBAM) and an enhanced decoder to better capture the complex features of sea ice.

\subsection{Benchmarking}
We further examine benchmarking frameworks in related fields, such as land cover classification and cloud detection, to understand how standardized evaluations have been successfully implemented in adjacent domains. These insights are critical for developing a robust benchmarking framework for sea ice type classification, enabling consistent and meaningful comparisons across different models.

In \cite{Papoutsis}, the authors present a comprehensive benchmark of 62 deep learning models for multi-label, multi-class land use land cover (LULC) image classification using the BigEarthNet \cite{BigEarthNet} dataset of Sentinel-2 satellite imagery. They have been motivated by a lack of reproducibility and comparability in the literature. With this benchmark, they assess traditional CNNs, Vision Transformers, and MLPs (Multi-Layer Perceptron), considering accuracy, training efficiency, and inference time. The authors proposed lightweight, scalable models based on Wide Residual Networks that outperformed ResNet50 by 4.5\% in F1-score with fewer parameters, aiming to facilitate the development of efficient deep learning architectures for remote sensing applications.

Similarly in \cite{López-Puigdollers}, the authors engage in a comprehensive benchmarking of deep learning models for cloud detection using satellite imagery from Landsat-8 and Sentinel-2. This work systematically contrasts these modern approaches against traditional operational algorithms. The findings indicate that while deep learning methods perform exceedingly well when tested within the same dataset, their effectiveness aligns closely with that of established methods when evaluated across disparate datasets or different sensor platforms. The paper underscores the significant influence of the training dataset on the performance of deep learning models, advocating for the establishment of standardized datasets from different sensors and robust benchmarking protocols to enhance future models for cloud detection.

Despite advancements in sea ice classification, a standardized benchmarking framework has yet to be established. While deep learning models have become the dominant approach for attempting to automate sea ice type classification, the absence of a dedicated model benchmark makes it challenging to systematically evaluate and compare their performance.
\begin{figure}[!ht]
    \centering
    \includegraphics[width=\linewidth ]{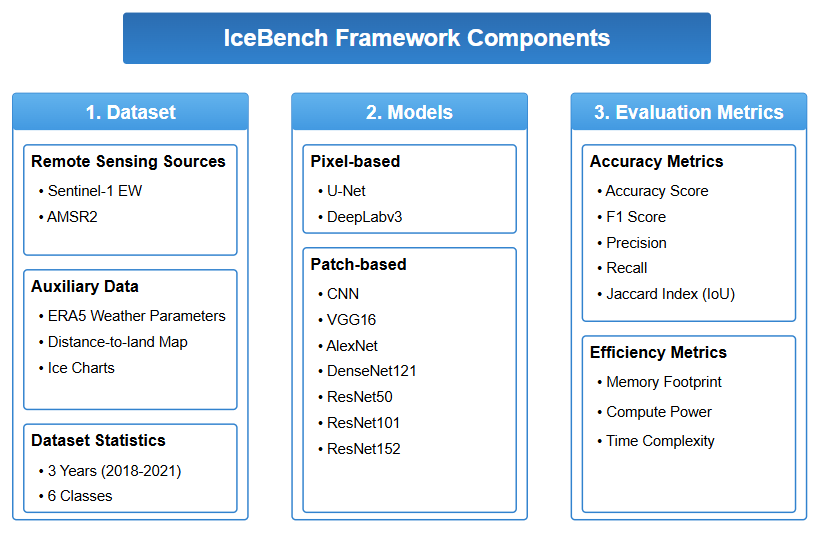} 
    \caption{\centering The Overall View of IceBench}
    \label{icebench}
\end{figure}
\section{IceBench}

In this section, we introduce IceBench, a benchmark framework for evaluating deep learning models in patch-based and pixel-based sea ice classification. As shown in Figure \ref{icebench}, IceBench comprises three key components: dataset, models, and evaluation metrics. Each component is selected based on current literature to address specific needs in sea ice type classification, making IceBench a robust tool for consistent and meaningful model comparisons.

\subsection{Dataset}
This section covers the candidate datasets considered for IceBench, the selected benchmark dataset, and the labeling process used for generating ground truth annotations.



\subsubsection{Candidate and Selected Dataset for IceBench}
A variety of datasets have been developed to support sea ice type classification, each offering unique advantages in terms of resolution, data modality, and geographical coverage. Table \ref{tab:dataset_comparison} summarizes the key features of each dataset, providing a comprehensive comparison of their variables, labels, spatial and temporal coverage, number of files, and resolution. Below, we briefly present each dataset:
\begin{table}[!ht]
\captionsetup{justification=centering}
\centering
\caption{Overview of candidate sea ice datasets.}
\label{tab:dataset_comparison}
\renewcommand{\arraystretch}{1.1}
\scriptsize 
\begin{tabular}{p{2.0cm} p{2.5cm} p{1.8cm} p{2.2cm} p{1.2cm} p{1.8cm}}
\toprule
\textbf{Dataset} & \textbf{Variables} & \textbf{Labels} & \textbf{Area and Time} & \textbf{\# Files} & \textbf{Pixel/Spatial Resolution} \\  
\midrule
AI4Arctic / ASIP Sea Ice Dataset (ASID-v2) \cite{ASIP} &  
Sentinel-1 EW (HH, HV), AMSR2, Incidence Angle, Geographical Data, Ice Charts from DMI &  
Provided as SIGRID code  &  
DMI charting areas (9 locations), 3/2018 - 5/2019 &  
461 NetCDF &  
Sentinel-1 and Ice Charts: 40 m; AMSR2: 2 km; Image Area: $400 \times 400$ km \\  
\midrule
AI4Arctic Challenge Dataset (RTT) \cite{ChallengeDataset} &  
Sentinel-1 EW (HH, HV), AMSR2, Incidence Angle, Geographical Data, ERA5 Weather Data, Ice Charts from DMI/CIS, Distance-to-land map &  
6 classes: OW, NI, YI, Thin FYI, Thick FYI, Old Ice &  
CIS/DMI regions (16 locations), 1/2018 - 12/2021 &  
533 NetCDF (513 train, 20 test) &  
Sentinel-1 and Ice Charts: 80 m; AMSR2: 2 km; Image Area: $400 \times 400$ km \\  
\midrule
AI4Arctic Challenge Dataset (Raw) \cite{ChallengeDataset} &  
Sentinel-1 EW (HH, HV), AMSR2, Incidence Angle, Geographical Data, ERA5 Weather Data, Ice Charts from DMI/CIS, Distance-to-land map &  
Provided as SIGRID code &  
CIS/DMI regions (16 locations), 1/2018 - 12/2021 &  
533 NetCDF (513 train, 20 test) &  
Sentinel-1 and Ice Charts: 40 m; AMSR2: 2 km; Image Area: $400 \times 400$ km \\  
\midrule
SI-STSAR-7 \cite{Wei2021} &  
Sentinel-1 A/B SAR (HH, HV)  &  
7 classes (OW, NI, GI, GWI, ThinFI, MedFI, ThickFI) &  
Hudson Bay, 10/2019 - 5/2020, 10/2020 - 4/2021 &  
164,564 samples &  
Image size: $32 \times 32$ pixels \\  
\midrule
S1 SAR-based sea ice cover \cite{Yiran} &  
 Sentinel-1 A/B SAR (EW mode) &  
Sea ice coverage &  
Arctic, 2019-2021 &  
2500/month &  
Resolution 40 m, Image area: $400 \times 400$ km \\  
\midrule
SAR-based Ice Types Dataset \cite{Khaleghian2020} &  
Sentinel-1A EW Level-1 GRD&  
5 classes (Water, Brash/Pancake, Young, Level FYI, Deformed) &  
North of Svalbard, Sep-Mar, 2015-2018 &  
19,029 samples&  
Resolution 40 m, Image sizes: $32 \times 32$ pixels \\  
\bottomrule
\end{tabular}
\end{table}

AI4Arctic / ASIP Sea Ice Dataset (ASID-v2) \cite{ASIP}: Produced through collaboration between the Danish Meteorological Institute (DMI), Technical University of Denmark (DTU), and Nansen Environmental Remote Sensing Center (NERSC), this dataset provides 461 netCDF files containing dual-polarized C-band Sentinel-1 EW images (HH, HV), Advanced Microwave Scanning Radiometer 2 (AMSR2) data, and manually created ice charts from DMI. It covers a time span from March 2018 to May 2019 and includes detailed geographical data and incidence angles, essential for accurate model training and sea ice analysis. Additionally, the dataset includes Sentinel-1 SAR data with ESA noise correction and Sentinel-1 SAR data with NERSC noise correction. This dataset, accessible through the DTU data portal, is tailored for Arctic sea ice monitoring and facilitates research on the variability of ice conditions across multiple seasons.

AI4Arctic Sea Ice Challenge Dataset \cite{ChallengeDataset}: Originally assembled for the AutoICE competition\cite{AutoIce}, this dataset comprises 533 files (513 for training and 20 for testing) that combine C-band Sentinel-1 SAR imagery, AMSR2 brightness temperatures, ERA5 weather data, and manual ice charts from the DMI and the Canadian Ice Service (CIS). Unique to this dataset is the availability of two versions: a raw version for detailed custom processing and a ready-to-train (RTT) version that simplifies the initial data preparation for model development. This dataset includes dual-polarized (HH, HV) Sentinel-1 Extra Wide Swath (EW) Level-1 Ground Range Detected (GRD) medium-resolution data along with AMSR2 passive microwave radiometer measurements \cite{Torres}. The Sentinel-1 SAR data processed with noise correction by the Nansen Environmental and Remote Sensing Center \cite{Korosov}. For the RTT version of the dataset, a polygon is assigned an ice type label if that type constitutes at least 65\% of the partial concentration within the polygon. 

SI-STSAR-7 Dataset \cite{Wei2021}: This labeled spatiotemporal dataset is based on 80 Sentinel-1 A/B SAR images, covering two freezing periods in Hudson Bay (October 2019 - May 2020 and October 2020 - April 2021). It includes seven sea ice classes: Open Water (OW), New Ice (NI), Grey Ice (GI), Grey White Ice (GWI), Thin First-Year Ice (ThinFI), Medium First-Year Ice (MedFI), and Thick First-Year Ice (ThickFI). Each data sample is a SAR image block containing dual-polarization data (HH and HV) with a spatial resolution of 32x32 pixels. Class labels are derived from weekly regional ice charts from the CIS, identifying regions where total ice concentration exceeds 90\% and the dominant ice type covers at least 90\% of the total ice area \cite{Song2021}. This dataset provides detailed temporal snapshots across freeze cycles, making it valuable for studying seasonal dynamics.

Sentinel-1 SAR-based Sea Ice Cover Dataset \cite{Yiran}:  Generated from Sentinel-1A and 1B SAR satellites in Extra Wide (EW) swath mode, this dataset covers the Arctic region from 2019 to 2021. The SAR data has a pixel size of 40x40 meters, and the derived sea ice cover product has a spatial resolution of 400 meters. Approximately 2500 SAR scenes are acquired each month, offering comprehensive monthly sea ice cover data for the Arctic. This dataset is provided in the NetCDF format, making it compatible with standard geospatial tools.

SAR-based Ice Types/Ice Edge Dataset for Deep Learning Analysis \cite{Khaleghian2020}: Designed for sea ice type and edge classification, this annotated dataset is based on 31 Sentinel-1A EW Level-1 Ground Range Detected (GRD) scenes, acquired over the region north of the Svalbard archipelago during winter months (September to March) from 2015 to 2018. With a spatial resolution of 40x40 meters, this dataset includes manually annotated polygons representing five sea ice classes: Water, Brash/Pancake Ice, Young Ice, Level First-Year Ice, and Deformed Ice. Additional Sentinel-2 optical imagery was used to support the manual annotations, while preprocessing with the European Space Agency’s Sentinel Application Platform (SNAP) software removed thermal noise. The dataset provides valuable labeled patches for understanding sea ice characteristics and edge transitions.

Among the datasets considered, the AI4Arctic Sea Ice Challenge Dataset \cite{ChallengeDataset} is a strong candidate for IceBench. It offers a substantial number of training files and integrates diverse data sources, including Sentinel-1 SAR imagery, AMSR2 brightness temperatures, ERA5 weather data, and ice charts from CIS/DMI. With extensive spatial and temporal coverage, it captures seasonal variations and regional differences in sea ice conditions, enabling models to learn from a wide range of environmental patterns. In conclusion, the raw version of this dataset provides a robust foundation for sea ice type classification, supporting the development of accurate and generalizable models.
\subsubsection{Labeling Process}
Ice charts, produced by national ice centers such as the United States National Ice Center (NIC), the Canadian Ice Service (CIS), and the Norwegian Meteorological Institute, provide essential data on sea ice conditions. These charts follow World Meteorological Organization (WMO) standards to ensure consistency and accuracy in sea ice monitoring. The WMO developed the egg code, a standardized system for encoding sea ice information by representing key ice parameters—such as concentration, stage of development, and form—within an oval diagram \cite{NSIDC}. Additionally, the Sea Ice Grid (SIGRID) format digitally encodes ice chart data into a machine-readable vector format, facilitating interoperability across different ice services and applications such as forecasting and navigation. Each ice chart divides a region into polygons representing homogeneous sea ice conditions, with multiple parameters assigned to each polygon. The key parameters include:
\setcounter{footnote}{0} 
\begin{figure}[!tb]
    \centering
    \includegraphics[width=4in , height = 3in]{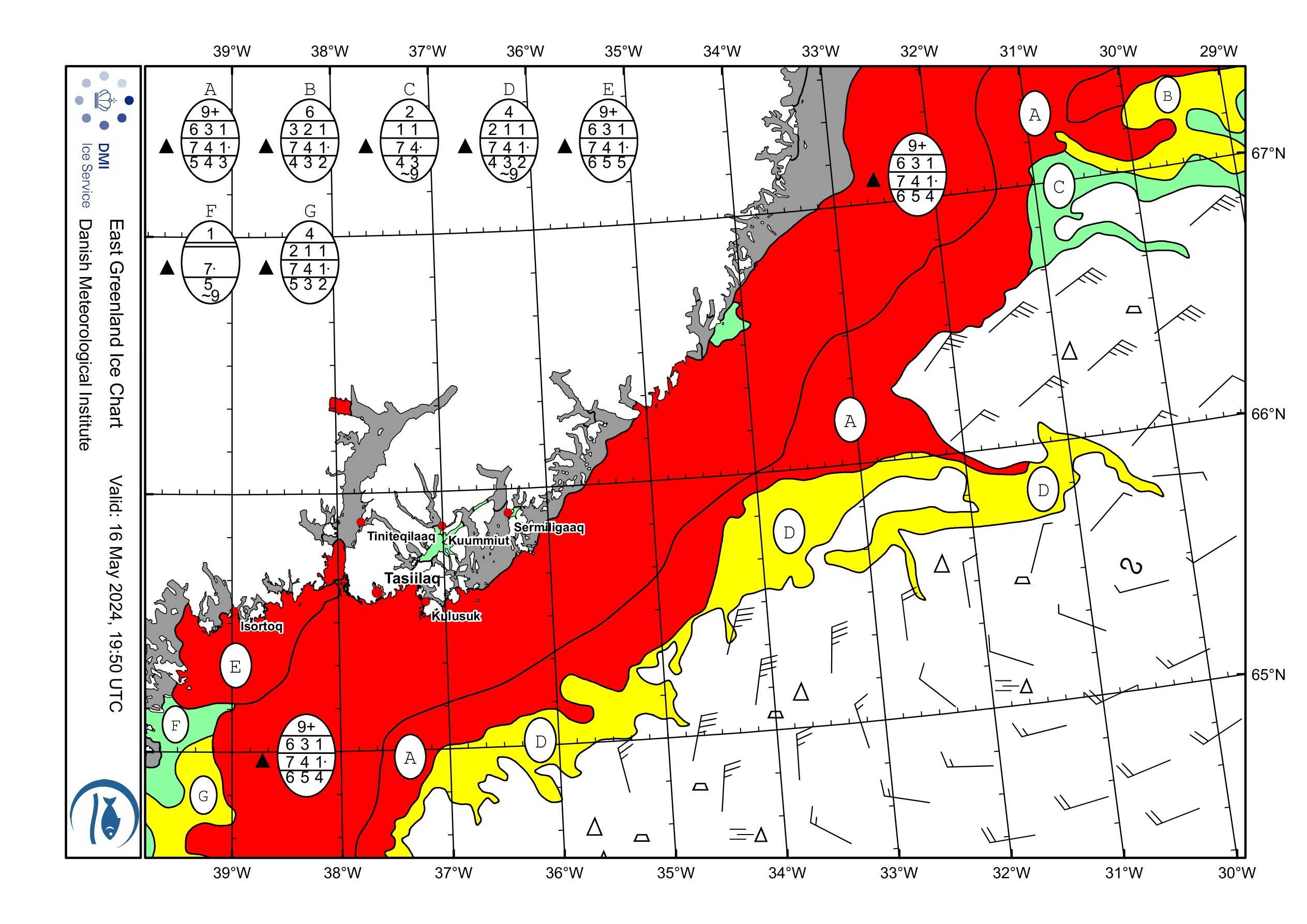} 
    \caption{ Example of an Ice Chart of Southeast Greenland, produced by the Greenland Ice Service at DMI, based on the egg code. Source:\protect\footnotemark.}
    \label{icechart}
\end{figure}
\footnotetext{\url{https://ocean.dmi.dk/arctic/icecharts.uk.php}}

\begin{itemize}
    \item Total Sea Ice Concentration (SIC): Indicates the percentage of sea ice coverage within each polygon, ranging from 0\% to 100\% (complete ice coverage). Partial sea ice concentration specifies the concentration levels of individual sea ice types within a polygon. Analysts assess partial concentrations for different ice types and assign values based on their observations.
    
    \item Stage of Development (SOD): Represents the type of sea ice, classifying it based on age and formation stage into categories such as new, young, first-year, multi-year, and old ice. SOD serves as an important indicator, as older ice is generally thicker and more stable. This classification is essential for understanding ice evolution, stability, and its impact on navigation and climate monitoring.
    
    \item Floe Size (FLOE): Refers to the size of the floating ice pieces within a polygon, with classifications ranging from small pancake ice to large floes. This parameter is particularly relevant for maritime navigation, as larger floes pose greater hazards to vessels.
\end{itemize}
Ice charts are predominantly created by ice analysts who manually annotate these parameters based on SAR and auxiliary data. Although ice charts offer high-quality annotations, they rely on the subjective judgment of analysts and limiting their use as precise ground truth data \cite{Cheng2020, Joint, ChallengeDataset}. Figure \ref{icechart} illustrates an example ice chart created by the Greenland Ice Service, which is annotated according to the WMO egg code standards.

Ice charts are commonly used as a reference for labeling SAR images in sea ice type classification. These ice charts provide polygon-level ice type information, including total ice concentration and the partial concentrations of up to three ice types. Researchers often simplify each polygon to its dominant ice type (e.g., if one type exceeds 65\% coverage). While efficient and rooted in standardized reporting, such polygon-level labels can miss spatial variations within each polygon. For finer-scale studies, researchers may generate their own labels at the pixel level by manually examining the imagery, a process that is far more time-consuming and demands expert knowledge. Although pixel-level labeling can capture subtler spatial details, it still relies on manual interpretation and thus introduces additional subjectivity. Consequently, whether researchers rely on existing ice chart labels or create new pixel-level labels, both approaches ultimately involve manual annotation. The choice typically depends on the desired level of detail and the available resources.


\subsection{Methods}
For IceBench to serve as a comprehensive benchmark for sea ice classification, we have carefully selected a \textit{representative} set of deep learning-based models, including both baseline and state-of-the-art methods. In this selection, we identified two distinct categories of sea ice type classification models, namely, patch-based classification models and pixel-based classification models.
Patch-based models classify larger image segments, capturing high-level patterns in sea ice, while pixel-based models provide fine-grained segmentation, delineating detailed spatial boundaries. To select representative models in each category, we conducted a thorough literature review, identifying dominant and widely used models proven effective for sea ice type classification. This ensures IceBench remains relevant and valuable to the research community. 

\subsubsection{Patch-Based Classification Models}

For patch-based classification, we selected several models that are particularly well-suited for capturing spatial hierarchies and extracting complex features. Key models include:
\setcounter{footnote}{0}
\begin{itemize}
\item \textit{CNN}\cite{Boulze}: CNNs excel in spatial hierarchy detection, making them effective for capturing broad structural patterns within sea ice imagery. Several studies have adapted and modified the CNN architecture to improve its performance specifically for sea ice type classification tasks \cite{Boulze, Li2017, Han, Han2020}. For our implementation, we utilized the publicly available GitHub repository, which corresponds to the implementation used in the work by Boulze et al \footnote{\url{https://github.com/nansencenter/s1_icetype_cnn}}

\item \textit{AlexNet} \cite{Krizhevsky}: AlexNet introduced ReLU activations, dropout for overfitting reduction, and GPU-based training, enabling efficient processing of large datasets. These innovations made AlexNet highly effective for extracting detailed features from high-resolution imagery, a critical requirement in sea ice classification. While AlexNet was originally designed for general image analysis tasks in \cite{Krizhevsky}, some studies have tailored it for sea ice classification with modifications to better handle SAR imagery \cite{Xu2017}.

\item \textit{VGG16}\cite{Simonyan}: VGG16 is a deep CNN architecture characterized by its simplicity and uniform design, using small 3x3 convolutional filters stacked sequentially to capture spatial hierarchies. This approach enables effective feature extraction while maintaining a manageable parameter size. While originally introduced in \cite{Simonyan} for general image classification, VGG16 has been employed in sea ice classification task \cite{Khaleghian2021}.

\item \textit{ResNet} \cite{He}: Known for its deep residual connections, ResNet mitigates gradient degradation which allows for stable training of very deep architectures. This feature makes it highly effective for detailed image analysis tasks. Originally introduced by \cite{He}, ResNet has since been adapted in several studies with task-specific modifications to enhance its performance for sea ice classification \cite{Song2018, Lyu2022, Zhang2021, Chen2023, Jiang2022}. In IceBench, we implement three versions of ResNet—ResNet-50, ResNet-101, and ResNet-152—offering a range of model complexities to assess performance across different computational constraints and classification challenges.

\item \textit{DenseNet121}\cite{Huang}: As a member of the DenseNet family, DenseNet121 consists of 121 layers with trainable weights and connects each layer to every other layer in a feed-forward manner. This architecture enhances feature reuse, reduces parameter counts, and optimizes efficiency and performance. Its design makes DenseNet particularly valuable in resource-constrained settings and well-suited for tasks requiring detailed feature extraction. Initially introduced for general image classification by \cite{Huang}, subsequent studies have adapted DenseNet for sea ice classification tasks, demonstrating its versatility and effectiveness in this domain \cite{Nagi, Kruk, Han2022}.

\end{itemize}

These models were originally introduced in their respective foundational papers and have since been extended or modified by various studies to address specific challenges in sea ice type classification. Where a GitHub link is cited, it indicates the implementation used; otherwise, the models were configured based on the original papers. Collectively, these architectures provide a diverse set of methods for patch-based classification, contributing distinct strengths in feature extraction, depth, and connectivity, ensuring a comprehensive evaluation across different scenarios for sea ice classification.

\subsubsection{Pixel-Based Classification Models}

For pixel-based classification, which involves segmenting images at the pixel level, we selected U-Net and DeepLabv3, two architectures specifically designed for segmentation and well-suited to the high-resolution requirements of sea ice analysis:
\setcounter{footnote}{0}
\begin{itemize}
\item \textit{U-Net}\cite{Ronneberger}: U-Net is a widely recognized architecture for tasks requiring precise pixel-level classification. Originally proposed by Ronneberger et al. \cite{Ronneberger} for biomedical image segmentation, U-Net has been successfully adapted for sea ice type classification in numerous studies \cite{Ren2020, Ren2021, Huang2021, Cantu, Ji2022, Wang}. For our implementation, we utilized the publicly available code from the \href{https://github.com/astokholm/AI4ArcticSeaIceChallenge}{AI4ArcticSeaIceChallenge}\footnote{\url{https://github.com/astokholm/AI4ArcticSeaIceChallenge}} repository. The U-Net model uses a series of convolutional layers with filter sizes [16, 32, 64, 64], a kernel size of (3,3), and a stride of (1,1), ensuring fine-grained feature extraction within its encoder-decoder architecture.

\item \textit{DeepLabv3}\cite{Lima2023}: DeepLabv3 leverages atrous (dilated) convolutions to effectively capture multi-scale contextual information, making it particularly suited for distinguishing between diverse sea ice types and accurately delineating their boundaries in satellite imagery. This architecture has been successfully applied to sea ice segmentation tasks in various studies \cite{Lima2023, Sun2023, Balasooriya, Zhang2022}. We used DeepLabv3 with a ResNet18 backbone, utilizing publicly available code at \href{https://github.com/geohai/sea-icebinary-ai4seaice}{sea-ice-binary-ai4seaice}\footnote{\url{https://github.com/geohai/sea-icebinary-ai4seaice}} repository. The model leverages an ASPP module for multi-scale feature extraction, supports flexible input channels, and refines outputs using bilinear interpolation.
\end{itemize}
These models were chosen based on their strong performance in pixel-level sea ice segmentation. Their accuracy make them ideal for addressing the challenges of sea ice segmentation, where fine-grained spatial detail is crucial. 

To evaluate implementation feasibility, we quantified model complexity using two key metrics: storage size (MB) and total trainable parameters (M). Table \ref{tab:model_complexity} compares the complexity of patch-based models and pixel-based models in IceBench. Patch-based models vary in efficiency, CNN-based architectures are lightweight and suited for fast inference, while VGG16 offers high capacity at a significant computational cost. DenseNet121 and ResNet variants strike a balance, with deeper architectures improving feature extraction but increasing resource demands. For pixel-based segmentation, U-Net is lightweight and efficient for real-time applications, whereas DeepLabV3, though slightly larger, enhances spatial detail extraction with advanced convolutions. This comparison underscores the importance of selecting the right model based on the specific application needs—lighter models are preferable for real-time inference and low-power environments, whereas larger models provide higher accuracy at the cost of increased computational and storage demands.

\begin{table}[H]
    \centering
    \caption{\centering Comparison of Model Complexity}
    \label{tab:model_complexity}
    \small
    \renewcommand{\arraystretch}{0.9} 
    \setlength{\tabcolsep}{3pt} 
    \begin{tabular}{l c c}
        \toprule
        \textbf{Model} & \textbf{Params (M)} & \textbf{Size (GB)} \\
        \midrule
        CNN        & 0.02   & 0.28  \\
        AlexNet    & 57.06  & 653   \\
        VGG16      & 134.29 & 1638  \\
        ResNet50   & 23.56  & 271   \\
        ResNet101  & 42.55  & 488   \\
        ResNet152  & 58.20  & 688   \\
        DenseNet121& 7.00   & 82    \\
        U-Net      & 0.52   & 4     \\
        DeepLabV3  & 1.43   & 11    \\
        \bottomrule
    \end{tabular}
\end{table}

\subsection{Metrics}
In this section, we present the evaluation metrics used to assess the performance of models in IceBench. Our framework includes both accuracy metrics, which measure classification and segmentation accuracy, and efficiency metrics, which assess computation resource requirements. These metrics allow for comprehensive evaluation of the capabilities of the models in various scenarios.

\subsubsection{Accuracy Metrics} We employ a mostly unified set of accuracy metrics applicable to both patch-based and pixel-based tasks. While some metrics, such as \textit{Intersection over Union (IoU)}, are more specific to pixel-based classification, others are common across both types of tasks. All metrics are computed using weighted averages, where class contributions are weighted by their respective pathc/pixel frequencies. 
\begin{itemize}
    \item \textit{Accuracy}: This is the most straightforward evaluation metric that measures the proportion of correct predictions (both true positives and true negatives) among all predictions:
    \begin{equation}
    Accuracy = \frac{TP + TN}{TP + TN + FP + FN}
    \label{eq:acc}
    \end{equation}
    where True Positives (\(TP\)) are instances correctly predicted as positive, True Negatives (\(TN)\) are instances correctly identified as negative, False Positives (\(FP\)) are instances incorrectly predicted as positive, and False Negatives (\(FN\)) are instances incorrectly predicted as negative.
    Accuracy is most effective when the dataset has a balanced number of positive and negative samples.

    \item \textit{Precision}: Precision reflects the proportion of true positive predictions among all positive predictions:
    \begin{equation}
    Precision = \frac{TP}{TP + FP}
    \label{eq:per}
    \end{equation}
    Precision is critical in scenarios where minimizing false positives is a priority.

    \item \textit{Recall}: Recall indicates the proportion of true positives identified among all actual positive instances:
    \begin{equation}
    Recall = \frac{TP}{TP + FN}
    \label{eq:re}
    \end{equation}
    Recall is essential to identify instances that might otherwise be overlooked.

    \item \textit{F1-Score}: The F1-Score is the harmonic mean of precision and recall, making it a reliable metric for evaluating models on imbalanced datasets where accuracy may be misleading. It is defined as:
    \begin{equation}
    F1 = 2 \times \frac{precision \times recall}{precision + recall}
    \label{eq:F1}
    \end{equation}
    In IceBench, the F1-score is a primary evaluation metric as it provides a balanced measure of model performance, ensuring a robust assessment in scenarios where class imbalances are common, such as sea ice classification.

    \item \textit{Intersection over Union (IoU)}: Also known as the \textit{Jaccard Index}, IoU quantifies the overlap between predicted and ground truth segmentation masks in pixel-based tasks:
     \begin{equation}
    Jaccard Index (IoU) = \frac{|A \cap B|}{|A \cup B|} = \frac{TP}{TP + FP + FN}
    \label{eq:iou}
    \end{equation}
    \(A\) and \(B\) represent the predicted and ground truth segmentation masks, while \(TP\), \(FP\), and \(FN\) denote true positive, false positive, and false negative pixels, respectively. The IoU metric is a key measure for evaluating segmentation quality, as it directly quantifies the overlap between the predicted and actual labels.
\end{itemize}


\subsubsection{Efficiency Metrics}
Within the IceBench framework, we consider three categories of efficiency metrics that capture memory footprint, use of compute power, and time complexity of the models. We use efficiency metrics to evaluate performance of the models during training and inference. In each category we leverage complementary metrics for comprehensive evaluation.  

\begin{itemize}
    \item \textit{Memory Footprint Metrics}: These metrics measure memory resources throughout the model lifecycle, for both training and inference phases:
    \begin{enumerate}
        \item Maximum Memory - Training (MaxMT): Maximum memory usage needed during the training process, measured in GB.
        \item Average Memory - Training (AvgMT): Average memory consumption during the entire training duration, measured in GB.
        \item Maximum Memory - Inference (MaxMI): Maximum memory usage needed during inference, measured in GB. 
        \item Average Memory - Inference (AvgMI): Average memory usage required during inference, measured in GB.
        
    \end{enumerate}
    Metrics for training and inference are chosen to provide detailed insights into resource usage across different phases of the model lifecycle.

    \item \textit{Compute Power Metrics}: These metrics quantify the consumption of compute resources, specifically CPU and GPU utilization. 
    \begin{enumerate}
       
        \item Total Power - Training (TotCT): total compute power used during training, measured in core hours.
        \item Total Power Inference (TotCI): total compute power used during inference, measured in core hours.
    
    \end{enumerate}
    The core hours for CPU and GPU usage are calculated as follows:
    \begin{equation}
    \begin{split}
    \text{Core Hours} &= \text{Usage Fraction} \times \text{Duration} \times \text{Computing Units}
    \end{split}
    \end{equation}
   
    In this formula, ``Usage'' (\%) represents the utilization of the CPU and GPU during the specified time interval. ``Duration'' (hours) refers to the total time for which the usage is measured. ``Computing Units'' indicates the number of CPUs or GPUs employed. This formula ensures that the calculated core hours accurately reflect the computational resources consumed during both training and inference phases.

    \item \textit{Time Complexity Metrics}: These metrics measure the time required for model operations:
    \begin{enumerate}
        \item Average Time per Epoch - Training (AvgET): Average time required to train one epoch, measured in minutes.
        \item Total Time - Training (TotTT): Total time required until convergence during training, measured in hours.
        \item Total Time - Inference (TotTI): Total time required to make a prediction, measured in minutes.
    \end{enumerate}
    The metrics provide a clear understanding of the temporal requirements for both training and inference.
    
\end{itemize}

By including these accuracy and efficiency metrics, IceBench enables a comprehensive evaluation of models with common metrics, including assessment of trade-offs between accuracy and efficiency. 

\section{Experimental Comparative Study}
This section presents a systematic model evaluation, outlining the experimental setup, including data processing, model selection, and validation. We then compare performance across multiple metrics to identify the most effective approaches.

\subsection{Experimental Methodology}
\begin{figure}[t]
    \centering
    \includegraphics[width=\columnwidth]{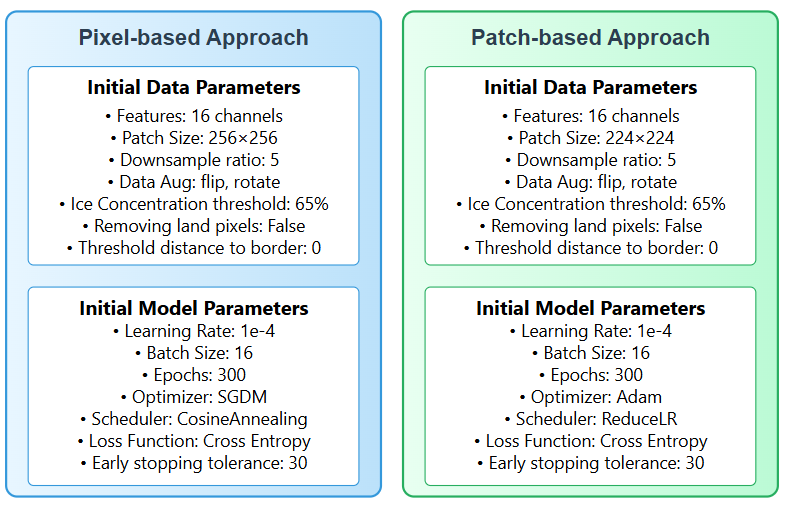}
    \caption{\centering Experimental Framework for Pixel-based and Patch-based Sea Ice Classification}
    \label{framework}
\end{figure}

Ensuring reliable and reproducible results in evaluation of models requires well-structured experimental design. Our approach to determine optimal data and model parameters drew from multiple sources, combining insights from extensive literature review for sea ice classification task, prior successful implementations, and the AutoICE Challenge results. Figure \ref{framework} provides an overview of the initial data and model parameters that form the foundation for our experimental pipeline detailed in the following subsections.

\setcounter{footnote}{0} 
\subsubsection{Data Processing: Feature Selection, Preprocessing, and Labeling}
Our experiments utilize the raw version of the AI4Arctic Sea Ice Challenge Dataset, which serves as the foundation for feature extraction and model training \cite{ChallengeDataset}.  The data are openly available at this link\footnote{\url{https://doi.org/10.11583/DTU.c.6244065.v2}}. This dataset includes 513 training files and 20 test files, and we consistently use the same training and test sets throughout our experiments. Based on insights from Chen et al. \cite{Chen2023} result on this dataset and related literature, we determined that model performance could be enhanced through the integration of diverse feature sets. To achieve this, we identified 16 features that encompass spatial, spectral, environmental, and temporal characteristics of sea ice. These features are summarized in Table \ref{tab:data_variables} and include inputs such as SAR imagery, brightness temperatures, meteorological parameters, and geographic and temporal data. For SAR-based features, we incorporated HH and HV polarizations alongside incidence angles, as they effectively represent sea ice properties. Additionally, distance maps and geospatial coordinates (longitude and latitude) were included to account for spatial variability in sea ice distribution. To distinguish between different ice types and open water, we leveraged passive microwave data from the AMSR2 instrument. Specifically, the 18.7 and 36.5 GHz frequencies for both horizontal and vertical polarizations were used, as these frequencies capture the spectral nuances of sea ice. We also included environmental variables that influence sea ice dynamics, such as wind components (eastward and northward at 10 meters), air temperature (at 2 meters), total column water vapor, and total column cloud liquid water. Since seasonal patterns play a crucial role, we added the month of image acquisition as a temporal variable to complete our feature set.


\begin{table}[ht]
\centering
\caption{\centering Key variables used for model training}
\label{tab:data_variables}
\footnotesize
\begin{tabular}{p{0.25\columnwidth} p{0.6\columnwidth}}
\hline
\textbf{Variable} & \textbf{Description} \\
\hline
nersc\_sar\_primary & Sentinel-1 image in HH polarization with NERSC noise correction \\
nersc\_sar\_secondary & Sentinel-1 image in HV polarization with NERSC noise correction \\
sar\_incidence\_angle & The incidence angle of the SAR sensor \\
distance\_map & Distance to land zones numbered with IDs ranging from 0 to 41 \\
sar\_grid\_latitude and sar\_grid\_longitude & Geographic information describing the common grid of the Sentinel-1 \\
btemp\_FFP & Brightness temperatures (Tb) for frequencies FF = [ 18.7, 36.5] and polarizations P = [v, h] \\
u10m\_rotated & ERA5 eastward component of the 10m wind rotated to account for the Sentinel-1 flight direction \\
v10m\_rotated & ERA5 northward component of the 10m wind rotated to account for the Sentinel-1 flight direction \\
t2m & ERA5 2m air temperature \\
tcwv & ERA5 total column water vapor \\
tclw & ERA5 total column cloud liquid water \\
month & Month of SAR data acquisition \\
\hline
\end{tabular}
\end{table}

Data preprocessing and preparation formed a critical foundation for our model training pipeline, encompassing multiple steps to ensure data quality and computational efficiency. At the highest level, our preprocessing workflow addressed three key challenges: feature alignment, computational optimization, and patch extraction strategies. The initial preprocessing phase focused on feature alignment, where we aligned all features with the Sentinel-1 SAR shape through resampling and interpolation, using a combination of averaging and max-pooling kernels specialized to different data types. Building upon this aligned dataset, we addressed the computational challenges of high-resolution data processing. After careful analysis, we recognized that using a downscale factor of zero (i.e., no downscaling) was computationally intensive and impractical for pixel-based models due to resource constraints. Therefore, following the findings of Chen et al. \cite{Chen2023}, we adopted a downscaling ratio of 5 that provided a reasonable trade-off between resolution and computational feasibility, ensuring that the spatial details necessary for accurate classification were preserved.

To ensure high-quality training data, our patch selection criteria, derived from extensive literature review on sea ice type classification, incorporated quality control measures. Our quality control approach focused on two primary considerations: the exclusion of land pixels and maintaining minimum distances from polygon borders to ensure patch purity. Distance from polygon borders was used to filter patches; patches close to polygon borders often contain mixed ice types, so we set a minimum distance threshold from borders, ensuring all pixels within a patch belong to a single ice type. 

The final phase of our data preprocessing pipeline involved implementing distinct patch extraction strategies for pixel-based and patch-based classifications. For pixel-based classification, we implemented dynamic random cropping with a $256\times256$ pixel size to expose models to different spatial regions, increasing data variability and improving generalization to diverse ice conditions. The epoch length was set to 500 steps for stable training. In contrast, patch-based classification requires a more structured approach. We systematically generated single-label patches across the entire training dataset, carefully maintaining patch purity by selecting regions where a single ice type was dominant. We implemented a systematic approach using fixed-size patches of $224\times224$ pixels with a stride of 100 pixels during the extraction process. For the patch-based approach the dataset comprising 23,144 training samples and 578 validation samples. These dimensions were carefully chosen to balance the capture of meaningful spatial patterns while maintaining computational efficiency.

Moreover, to increase the robustness and generalizability of our models, we applied data augmentation techniques. Data augmentation helps with preventing overfitting by artificially expanding the training dataset and introducing variability. We applied transformations such as rotation by up to ±10 degrees to simulate different viewing perspectives, and vertical flipping to introduce mirror images. These augmentations mimic real-world variations and help the model become invariant to such changes.

\setcounter{footnote}{0} 
We utilized the ice charts provided in the dataset for labeling purposes. While previous studies commonly employ a 50\% threshold to determine dominant ice type, we used a more conservative approach. After normalizing the partial concentrations by the total SIC, we established a 65\% threshold for identifying dominant ice types. This higher threshold significantly reduces labeling ambiguity and enhances the model's ability to distinguish between different ice classes. For practical implementation, we grouped similar SODs into broader categories, resulting in a simplified but meaningful set of six ice type classes. For details on the grouped codes and classes, refer to Table \ref{tab:ice_classes_wmo}. To ensure consistent and efficient label assignment, we leveraged the AutoICE challenge starter pack \href{https://github.com/astokholm/AI4ArcticSeaIceChallenge}{AI4ArcticSeaIceChallenge}\footnote{\url{https://github.com/astokholm/AI4ArcticSeaIceChallenge}} repository, which automates the process of translating ice chart annotations into defined class structure.

\begin{table}[htbp]
\centering
\caption{\centering Sea Ice Type Class Names with Corresponding Labels and Codes}
\label{tab:ice_classes_wmo}
\begin{tabular}{c c l}
\hline
\textbf{Label} & \textbf{SIGRID-3 Code} & \textbf{Class Name} \\ \hline
0 & 0, 80 & Open Water \\
1 & 81, 82 & New Ice \\
2 & 83–85 & Young Ice \\
3 & 87–89 & Thin First-Year Ice (FYI) \\
4 & 86, 91, 93 & Thick First-Year Ice (FYI) \\
5 & 95–97 & Old Ice (moer than 1 Year) \\
\hline
\end{tabular}
\end{table}

\subsubsection{Model Parameter Selection and Validation Strategy}
Figure \ref{framework} provides an overview of the initial model parameters. We selected distinct model parameters for each approach, considering their fundamental architectural differences. Our pixel-based models were initialized with parameters from Chen et al. \cite{Chen2023}, whose approach achieved the highest performance in the AutoICE Challenge. This included network architectures, hyperparameters, learning rates, batch sizes, and optimization algorithms that were proven effective in their experiments. For the patch-based models, we conducted a thorough literature review to identify optimal ranges for model parameters. We considered best practices and successful configurations from recent studies in the field \cite{Boulze,Kruk, Song2018}. Parameters such as patch size, stride, network depth, and activation functions were selected based on their effectiveness in similar image classification tasks. We also took into account the general impact of these parameters across different models to ensure that our selections were robust and widely applicable. The chosen parameters were fine-tuned to suit the specifics of our dataset and classification objectives.

For model validation, we used a fixed validation set of 18 randomly selected from train files, ensuring consistent evaluation during training and parameter tuning. Early stopping was applied with a patience of 30 epochs to prevent overfitting. Final model evaluation was conducted using the AutoICE Challenge test set, allowing direct performance comparison with other approaches in the field. The evaluation methodology was tailored to each model type. The testing methodology was adapted to accommodate each model's architectural requirements. The pixel-based models processed the entire test files, preserving spatial continuity and mimicking real-world deployment conditions. In contrast, the patch-based models evaluated single-label patches extracted from the test files, matching the dimensions used during training. This approach maintained methodological consistency while allowing us to apply standard performance metrics.

\subsubsection{Experimental Setup}
For our experimental setup, we utilized the PyTorch framework, known for its flexibility and efficiency in deep learning research. The experiments were conducted on a system equipped with an NVIDIA RTX A6000 GPU and an Intel Xeon Silver 4310 CPU (2.10 GHz base, 48 cores) with 62 GB of RAM. The code used to implement the models and conduct the experiments described in this paper is available at the \href{https://github.com/bdlab-ucd/IceBench}{IceBench github repository}.

\subsection{Experimental Results }
We conducted a comprehensive evaluation of models within the IceBench framework. Our evaluation began with a systematic assessment of all models using standardized metrics and testing protocols. For each model, we computed accuracy metrics as well as computational efficiency metrics. We identified the top-performing models within each classification category based on the F1-score metric. The final phase of our analysis involved a thorough comparison between the leading models from both approaches, examining their relative strengths, limitations, and performance trade-offs.

\begin{figure}[tb]
    \centering
    \captionsetup{justification=centering} 
    \begin{minipage}{\textwidth}
        \subfloat[]{
            \includegraphics[width=0.48\textwidth]{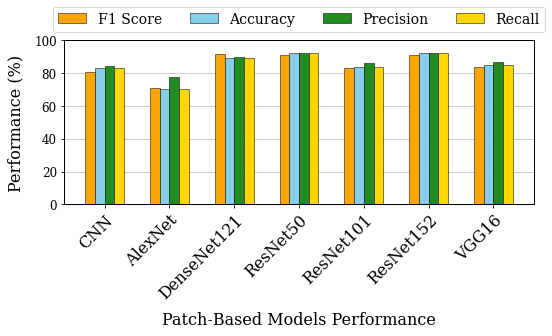}
            \label{fig:Classification_findbase}
        }
        \hfill
        \subfloat[]{
            \includegraphics[width=0.48\textwidth]{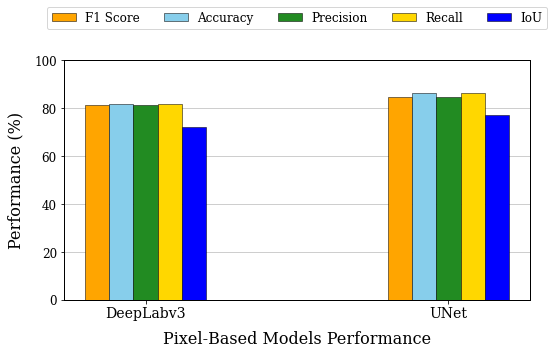}
            \label{fig:Segmentation_findbase}
        }
        \caption{Performance comparison of (\textbf{a}) patch-based and (\textbf{b}) pixel-based models on the test set.}
        \label{fig:performance_comparison}
    \end{minipage}
\end{figure}

Beginning with patch-based classification, our analysis encompassed various architectures as shown in Figure \ref{fig:Classification_findbase}. DenseNet121 achieves the highest F1 score at 91.57\%, while ResNet152 demonstrates a very similar F1 score. Meanwhile, ResNet50 and ResNet152 achieve the highest accuracy at 92.22\%. While these metrics are significantly better than simpler architectures like AlexNet (F1: 70.87\%) and basic CNN (F1: 80.53\%), VGG16 falls behind the top performers like DenseNet121 and ResNet variants. This middling performance could be attributed to VGG16's relatively simple architecture that uses repeated blocks of convolutional layers with small filters. While this design is effective for many computer vision tasks, it lacks the advanced features like skip connections (ResNet) or dense connectivity (DenseNet) that help modern architectures achieve superior performance in complex tasks like sea ice classification. When training DenseNet121 from scratch instead of using ImageNet weights \cite{Imagenet}, the F1-score dropped significantly to 78.29\%, highlighting advantage of transfer learning. Based on these compelling results, we selected DenseNet121 with ImageNet pretrained weights (hereafter referred to simply as DenseNet) as our top-performing patch-based model \cite{Russakovsky2015}. 
Similarly, for the pixel-based segmentation approach, we focused on pixel-based classification architectures, specifically evaluating U-Net and DeepLabV3 models. As shown in Figure \ref{fig:Segmentation_findbase}, U-Net achieves higher scores in F1 (84.78\% vs 81.32\%), accuracy (86.36\% vs 81.80\%), precision (84.68\% vs 81.59\%), recall (86.36\% vs 81.80\%), and IoU (77.18\% vs 72.03\%). U-Net's better performance can be attributed to its symmetric encoder-decoder architecture with skip connections, which is particularly effective for sea ice pixel-base classificaiotn as it preserves both fine-grained spatial details and global context information. The skip connections allow the network to maintain high-resolution features from the encoder path, which is crucial for accurate ice type boundary delineation. While DeepLabv3 also shows solid performance above 80\% in most metrics, its slightly lower performance might be due to its atrous convolution approach, which, although effective for general pixel-base classificaiotn, may not be as optimal for capturing the specific texture and boundary patterns characteristic of different sea ice types. Both models achieve relatively high IoU scores (over 70\%), indicating good overlap between predicted and ground truth segmentations, with U-Net's higher IoU of 77.18\% suggesting more precise boundary predictions.

These results suggest that while both approaches can achieve strong performance in sea ice type classification, the choice between them may depend on specific operational requirements and computational constraints.

While our accuracy metrics provided insights into classification capabilities, we complemented this with efficiency metrics focused on practical deployment considerations. Table \ref{tab:efficiency_metrics} presents these efficiency metrics across all models, revealing distinct patterns between pixel-based and patch-based approaches. Pixel-based and patch-based approaches exhibit distinct computational footprints, driven by how each method processes images, their respective epoch lengths, and overall convergence pattern. Resource utilization patterns reveal notable trends across approaches, leveraging 48 CPUs and 1 GPU for both training and inference.


Pixel-based models require higher core-hour consumption due to slower convergence and full-resolution image processing during validation and inference. Their smaller epoch length enables each epoch to process less data, resulting in lower memory usage per training step and faster per-epoch computation. However, they require more epochs to achieve convergence compared to patch-based models, leading to substantially longer total computation times (27.5–38.3 hours) than their patch-based counterparts (13.8–19.0 hours). Additionally, their computationally intensive inference phase, which involves processing and labeling each pixel in full-resolution images, results in longer inference times of approximately four minutes. 
In contrast, patch-based models exhibit higher memory consumption and significantly longer epoch durations due to their larger epoch length, which is five times larger compare to pixel-based models. As a result, these models have a longer epoch duration (about 12 minutes per epoch) and a greater number of iterations, with total training times ranging from approximately 6.3 to 7.5 hours. Additionally, the smaller fixed patch size used during inference facilitates considerably faster inference (around 0.7 minutes total), as less data is processed at once. Among the patch-based models, the ResNet family exhibit similar performance profiles, with minimal differences in computational efficiency. DenseNet offers a balanced performance profile with moderate memory usage during training. However, VGG16 records the highest memory spike during training, making it less ideal for resource-constrained environments.
Overall, pixel-based models preserve spatial relationships but are computationally intensive, while patch-based models offer faster inference, making them ideal for real-time applications. The choice depends on priorities: training efficiency, inference speed, or memory constraints.
\begin{table}[tb]
\renewcommand{\arraystretch}{1.05}
\setlength{\tabcolsep}{2pt} 
\centering
\caption{Comparison of Model Efficiency Metrics for Pixel-based and Patch-based Models}
\label{tab:efficiency_metrics}
\fontsize{8pt}{9pt}\selectfont 
\begin{tabular}{@{}l c c c c c c c c c@{}}
\hline
\multirow{2}{*}{\textbf{Model}} &
\multicolumn{5}{c}{\textbf{Training Phase}} &
\multicolumn{4}{c}{\textbf{Inference Phase}} \\
\cline{2-10}
& \begin{tabular}[c]{@{}c@{}}AvgMT\\ (GB)\end{tabular} &
\begin{tabular}[c]{@{}c@{}}MaxMT\\ (GB)\end{tabular} &
\begin{tabular}[c]{@{}c@{}}TotCT\\ (Hrs)\end{tabular} &
\begin{tabular}[c]{@{}c@{}}AvgET\\ (Min)\end{tabular} &
\begin{tabular}[c]{@{}c@{}}TotTT\\ (Hrs)\end{tabular} &
\begin{tabular}[c]{@{}c@{}}AvgMI\\ (GB)\end{tabular} &
\begin{tabular}[c]{@{}c@{}}MaxMI\\ (GB)\end{tabular} &
\begin{tabular}[c]{@{}c@{}}TotCI\\ (Hrs)\end{tabular} &
\begin{tabular}[c]{@{}c@{}}TotTI\\ (Min)\end{tabular} \\
\hline
\multicolumn{10}{l}{\scriptsize\textbf{Pixel-based Models}} \\
U-Net          &2.3 & 3.4&38.3 &4.6 & 7.2& 3.8 & 4.1 & 0.03& 4.4\\
DeepLabv3      &2.2 & 3.3 & 27.5 &4.6& 5.5 & 3.7 & 4.1 &0.03 & 4.0 \\
\multicolumn{10}{l}{\scriptsize\textbf{Patch-based Models}} \\
CNN            &4.1 &5.1 &13.8&12.3 & 6.5 &4.9 &4.9 &0.01 & 0.71 \\
AlexNet        &5.0 & 5.2& 15.7 &12.3 &7.1 &5.0 &5.0 &0.01 & 0.70\\
DenseNet121    &4.2 &4.4 &17.9 & 12.6& 7.2 &4.9 &4.9 & 0.01 & 0.73\\
ResNet50       &5.1&5.5 &15.1&12.4 &6.3 & 4.9 & 4.9&0.01& 0.74\\
ResNet101      & 4.3& 5.1&18.6 &12.5 &7.4 &5.0 &5.0 &0.01 &0.74\\
ResNet152      &4.3&5.2 &17.4&12.4 &6.3 & 5.0 & 5.0&0.01& 0.75\\
VGG16          &3.8 & 5.9&19.0 &12.4 & 7.5 &4.6 &4.6 &0.01&0.73 \\
\hline
\end{tabular}
\end{table}

Based on the accuracy metrics, we identified U-Net and DenseNet as the top performers in pixel-based and patch-based categories, respectively. The key challenge in this comparison arises from their distinct classification granularities: U-Net generates pixel-wise predictions, while DenseNet assigns a single classification to entire patches. 
To ensure a fair comparison between pixel-based and patch-based models, we established a uniform evaluation methodology using pixel-level ground truth as the reference standard. Test images were processed by both model types, and the patch-based DenseNet's predictions were mapped back to the pixel level to align with the ground truth for direct comparison. Table \ref{tab:performance_metrics_like_seg} presents these comparative results, revealing a significant performance disparity. The patch-based DenseNet showed markedly lower performance metrics when evaluated against pixel-level ground truth, primarily due to its inability to handle mixed-label scenarios. When patches contain pixels from multiple ice classes, a common occurrence in real-world sea ice imagery, the model must make a single classification decision for the entire patch, inevitably compromising pixel-level accuracy.

This direct comparison between the leading models yields crucial insights into the practical implications of model selection for sea ice type classification tasks. While patch-based approaches offer computational efficiency and good performance for homogeneous regions, their accuracy decreases significantly when precise pixel-level classifications are required, particularly in areas with diverse ice types. These findings underscore the importance of aligning model selection with specific application requirements - choosing pixel-based approaches for applications requiring high spatial precision, and patch-based approaches for scenarios where computational efficiency and broader ice type characterization are prioritized.

\begin{table}[ht]
\centering
\caption{Comparative Performance Metrics of U-Net and DenseNet for Fair Evaluation}
\label{tab:performance_metrics_like_seg} 
\begin{tabular}{l c c c c c}
\hline
Model         & F1-score & Accuracy & Precision & Recall & IoU \\\hline
U-Net         & 84.78    & 86.36    & 84.68    & 86.36  & 77.18    \\
Densenet121    & 53.89    & 54.77    & 63.79     & 54.77  & 38.51     \\
\hline
\end{tabular}
\end{table}
\section{Case Studies with IceBench} 
\begin{figure}[ht]
    \centering
    \includegraphics[width=\linewidth]{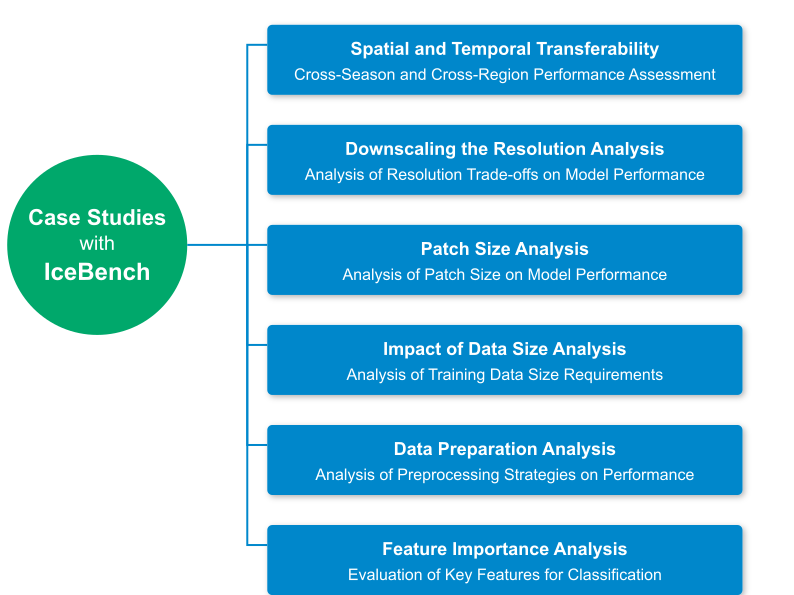} 
    \caption{\centering Key Investigations in Sea Ice Type Classification Using IceBench }
    \label{case_studies}
\end{figure}

Understanding the sensitivity of sea ice type classification models to various parameters is essential for building robust, operational systems. Therefore, in this section, we use IceBench to investigate critical factors influencing model performance in real-world scenarios. Figure \ref{case_studies} outlines the key research questions that guide our analysis.  
Toward this end, we addressed six key research questions:  
\begin{enumerate}
    \item How well do the models generalize across different temporal and spatial domains? We evaluate the models' ability to transfer knowledge to unseen seasons and geographical areas.
    \item What is the optimal balance between spatial context and computational efficiency? We investigate the impact of image downscaling ratios on model performance.
    \item How does patch size affect model performance? We analyze the trade-offs between patch dimensions and computational resources.
    \item What is the minimum training data size needed for robust performance? We examine the relationship between dataset volume and model accuracy.
    \item How do different data preparation strategies affect model performance? We assess the impact of preprocessing approaches and land masking methods.
    \item Which input channels are most critical for classification accuracy? We identify the most influential spectral and auxiliary channels for effective classification.
\end{enumerate}
 
Before addressing our research questions in these case studies, we first validated the effectiveness of our hyperparameter optimization process to ensure that the model configurations are well-calibrated and provide a reliable baseline for subsequent analyses. Our investigation encompassed both patch-based and pixel-based classification models, focusing on key parameters that significantly influence model performance: learning rate, batch size, optimizer choice, scheduler configuration, and the number of U-Net layers.

Our initial experiment focused on two key hyperparameters: learning rate and batch size. The learning rate dictates the step size for updating weights, directly influencing convergence speed and stability. Batch size determines the number of samples processed per update, balancing training efficiency and stability. 

\begin{figure}[!htbp]
    \centering
    \captionsetup{justification=centering} 
    \begin{minipage}{\textwidth}
        \subfloat[]{
            \includegraphics[width=0.48\textwidth]{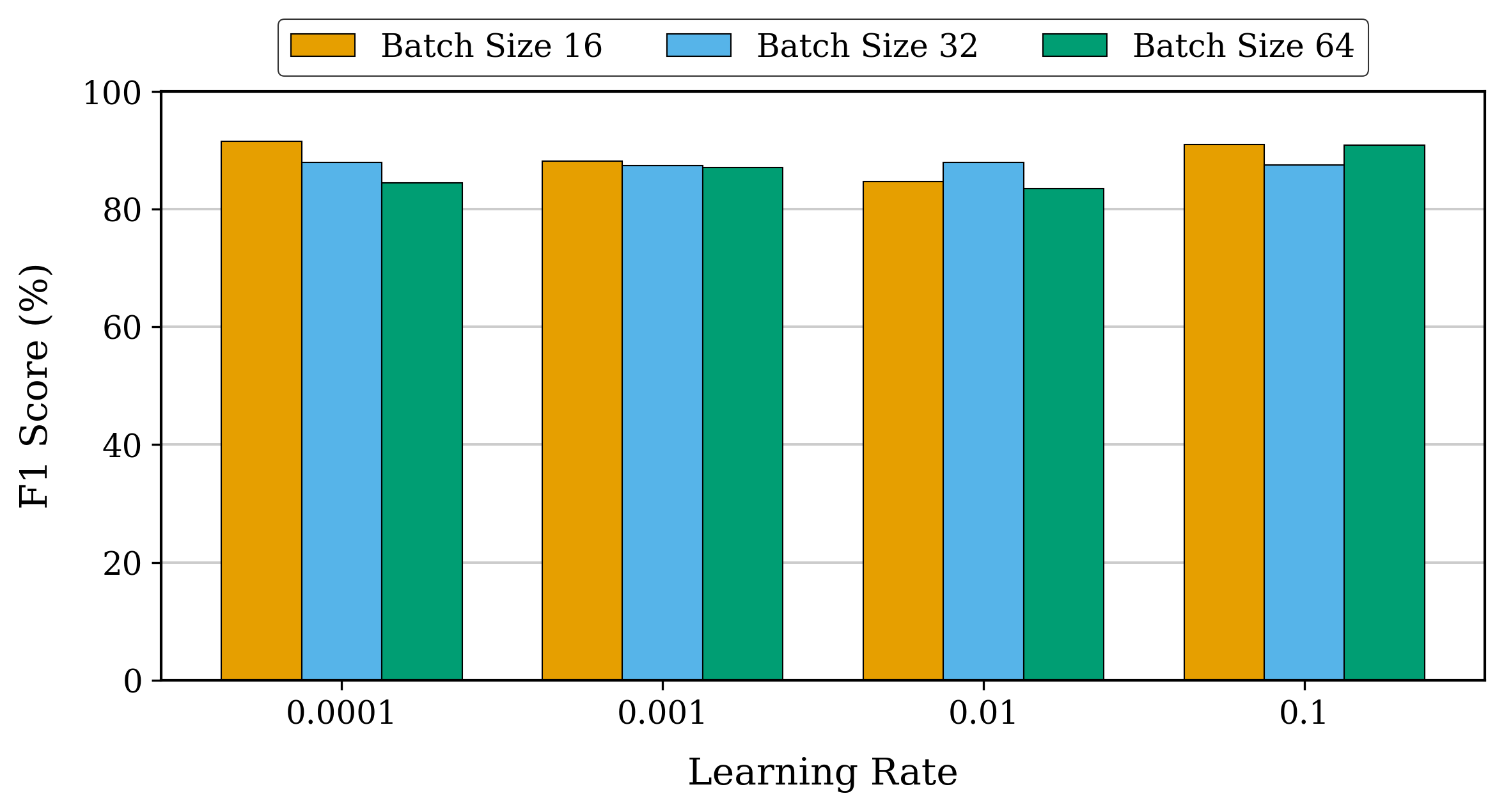}
            \label{fig:Clf_behavioral}
        }
        \hfill
        \subfloat[]{
            \includegraphics[width=0.48\textwidth]{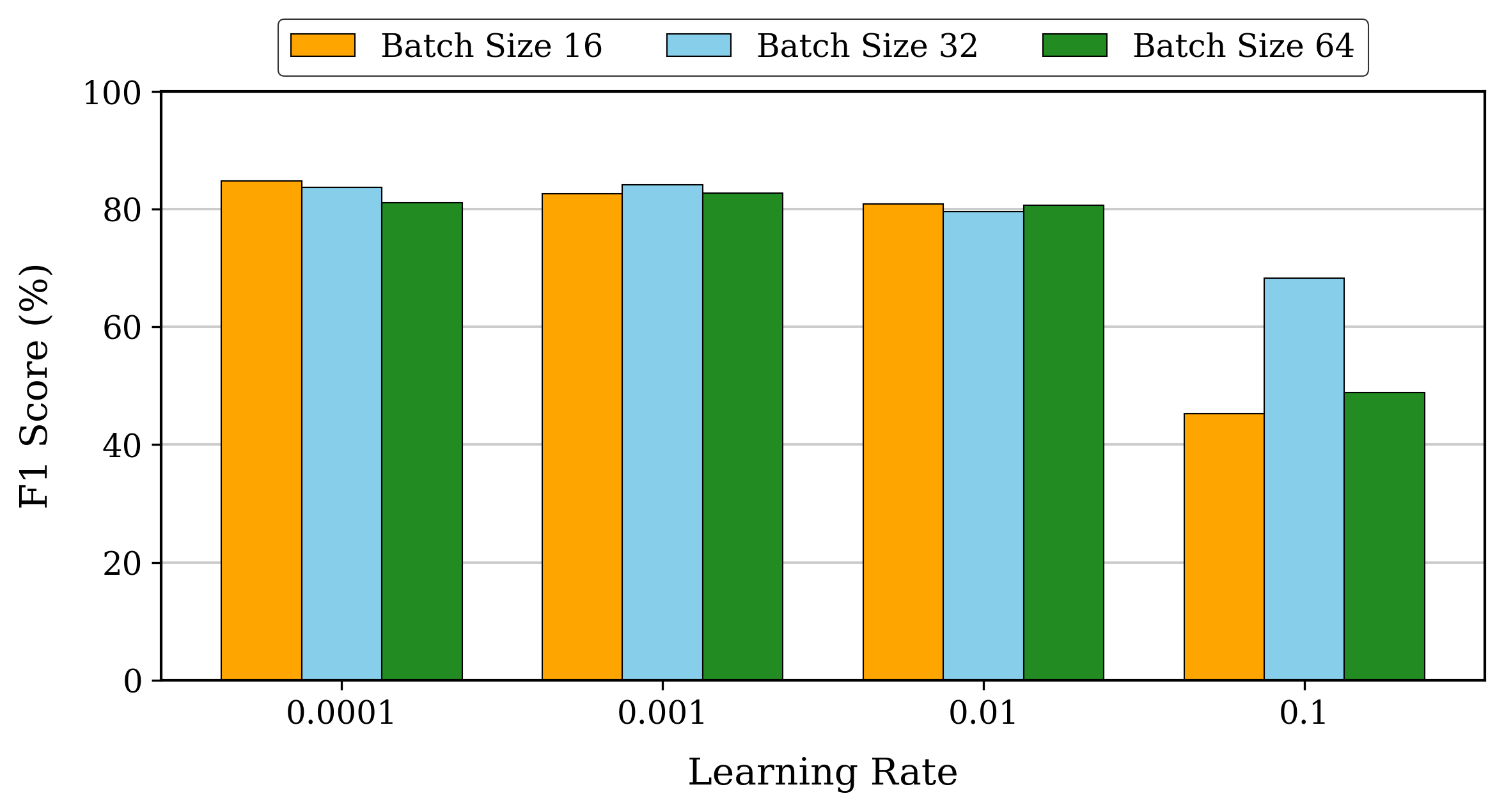}
            \label{fig:Seg_behavioral}
        }
        \caption{Sensitivity analysis of (\textbf{a}) DenseNet and (\textbf{b}) U-Net. The effect of learning rate and batch size on performance.}
        \label{fig:sensitivity_analysis}
    \end{minipage}
\end{figure}

For DenseNet, we tested different learning rates and batch sizes, as shown in Figure \ref{fig:Clf_behavioral}.  Using the Adam optimizer with a Reduce-on-Plateau scheduler as our baseline, we tested learning rates from 0.0001 to 0.1 and batch sizes of 16, 32, and 64, evaluating their impact on training performance based on F1-score. At a conservative learning rate of 0.0001, the model demonstrated robust performance across all batch sizes, with batch size 16 showing slightly superior results. When increasing the learning rate to 0.001, we observed a minor degradation in F1-scores across batch sizes, though the model maintained stable performance. Interestingly, at a higher learning rate of 0.1, the model showed remarkable resilience, achieving an F1-score of 91.00\% with batch size 16 and even surpassing lower learning rate configurations in accuracy and precision. At this higher learning rate, larger batch sizes also showed promise, with batch size 64 achieving a strong F1-score of 90.85

Expanding our analysis to pixel-based segmentation, we conducted a similar evaluation for the U-Net model. Figure \ref{fig:Seg_behavioral} presents the F1-scores across different learning rates (0.0001 to 0.1) and batch sizes (16, 32, and 64), using SGD optimizer with a Cosine Annealing scheduler as the baseline configuration. U-net demonstrated distinct behavior patterns across different parameter combinations. At a lower learning rate of 0.0001, we observed consistently superior performance, particularly with a batch size of 16, achieving an F1-score of 84.78\%, accuracy of 86.36\%, and a Jaccard Index of 77.18\%. As we increased the batch size to 64 under this learning rate, the performance showed slight degradation while maintaining overall robustness. When testing higher learning rates of 0.001 and 0.01, the model maintained strong performance, especially with smaller batch sizes of 16 and 32. However, at a learning rate of 0.1, we observed a significant drop in F1-scores across all batch sizes, indicating this learning rate exceeded the optimal range for stable training. Smaller batch sizes, especially batch size 16, consistently achieve higher F1-scores due to better gradient estimates and more frequent updates. In summary, lower learning rates (0.0001 and 0.001) with smaller batch sizes (16 and 32) yield the better F1-scores. 

\begin{figure}[!htbp]
    \centering
    \begin{minipage}{\textwidth}
        \subfloat[]{
            \includegraphics[width=0.48\textwidth]{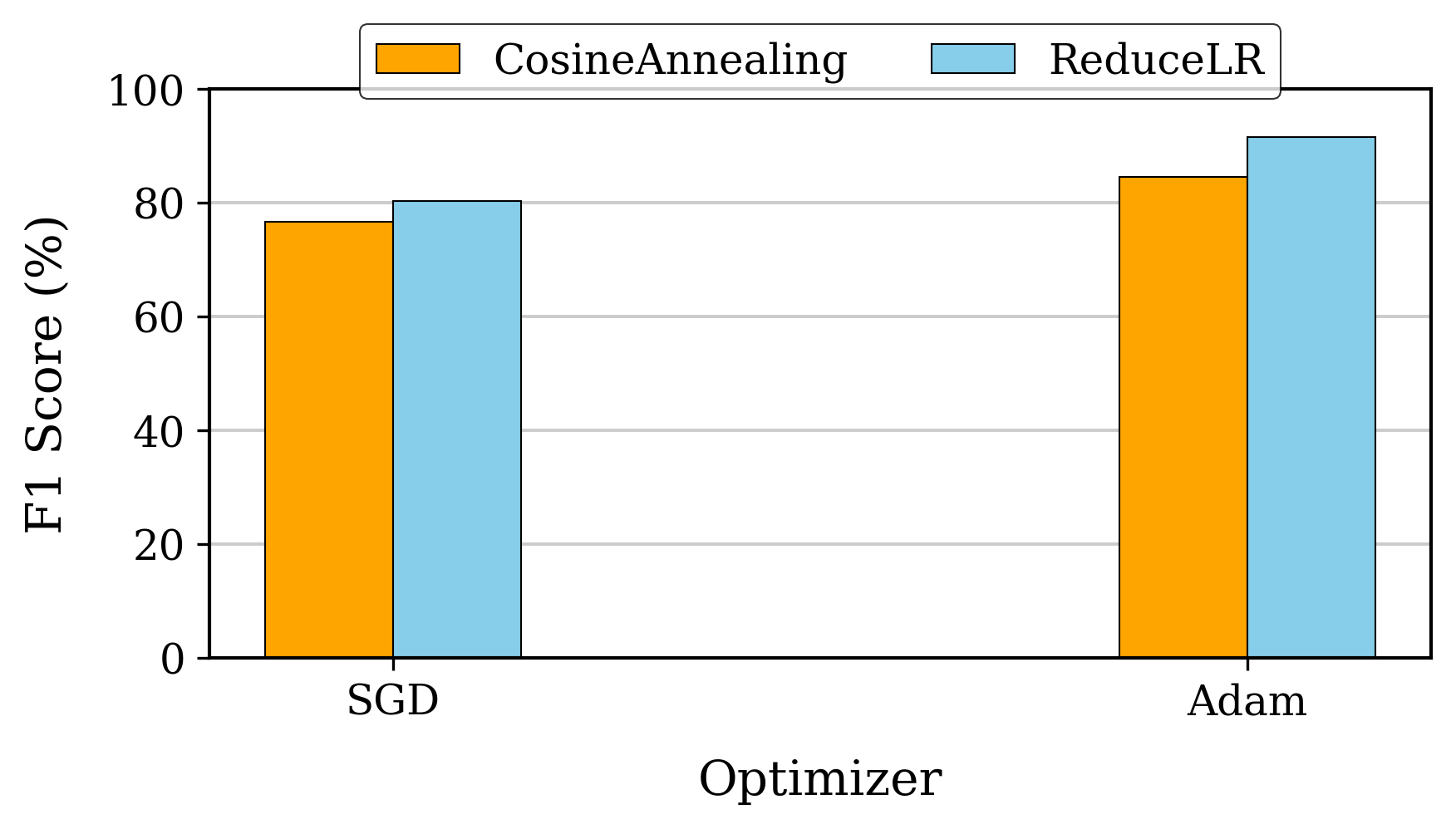}
            \label{fig:Clf_Opt}
        }
        \hfill
        \subfloat[]{
            \includegraphics[width=0.48\textwidth]{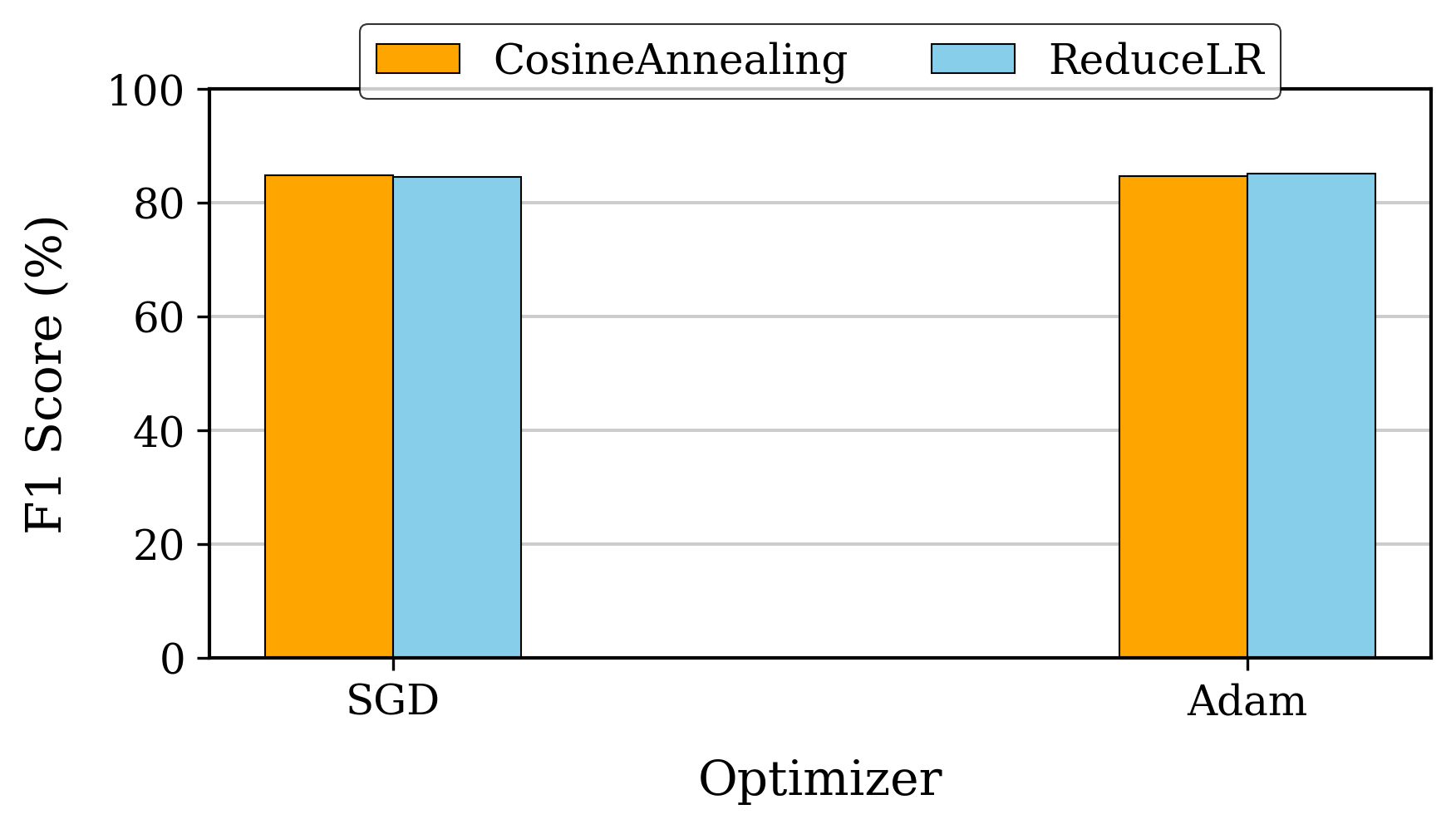}
            \label{fig:Seg_Opt}
        }
        \caption{Sensitivity analysis of (\textbf{a}) DenseNet and (\textbf{b}) U-Net: The effect of optimizer and learning rate scheduler on performance.}
        \label{fig:optimizer_scheduler}
    \end{minipage}
\end{figure}

Building on our hyperparameter analysis, we next examined the impact of optimizer choice and scheduler configuration on model performance. These factors play a crucial role in training dynamics—optimizers influence weight updates and convergence speed, while schedulers adjust learning rates to enhance stability and prevent overfitting.
For DenseNet, we compared the performance of Adam and SGD optimizers, each paired with either a Reduce-on-Plateau or Cosine Annealing scheduler. Figure \ref{fig:Clf_Opt} illustrates the F1-scores across these configurations, highlighting their effects on training efficiency and classification accuracy. The combination of Adam optimizer with Reduce on Plateau scheduler emerged as the optimal choice, achieving our highest F1-score of 91.57\%. This comprehensive analysis revealed that for patch-based classification, a learning rate of 0.0001 combined with a batch size of 16 provides the best balance for effective learning, while demonstrating the model's robustness across a wide range of parameter settings.

Similarly, for U-Net, we evaluated the impact of optimizer and scheduler selection on segmentation performance. As shown in Figure \ref{fig:Seg_Opt}, we tested SGD and Adam optimizers alongside Cosine Annealing and Reduce-on-Plateau schedulers, analyzing their influence on F1-scores and convergence behavior. The combination of Adam optimizer with Reduce on Plateau scheduler proved most effective, achieving an F1-score of 85.03\% with batch size 16. This superior performance of smaller batch sizes, particularly batch size 16, can be attributed to more frequent model updates and better gradient estimates.

Continuing our investigation, we analyzed the impact of model architecture variations on performance, specifically focusing on the depth of U-Net. Our experiments compared U-Net variants with 4 and 5 encoding/decoding layers. The 5-layer configuration (with layer depths of 32, 32, 64, 64, 128) achieved an F1-score of 82.63\%, accuracy of 83.60\%, and a Jaccard Index of 73.54\%. Interestingly, the 4-layer architecture [32, 32, 64, 64] demonstrated superior performance with an F1-score of 84.78\%, accuracy of 86.36\%, and a Jaccard Index of 77.18\%. This finding suggests that increasing model complexity beyond four layers does not necessarily yield better results for our specific task.

Throughout these experiments, our initial parameter selections consistently demonstrated strong performance, validating our preliminary choices for the sea ice type classification task. 


\subsection{Spatial and Temporal Transferability} 
In sea ice monitoring, a model's ability to generalize across different locations and seasons is essential for practical deployment. This transferability determines whether a model trained on data from one location or season can maintain its performance when applied to another. 

First, we define the seasonal classifications and categorize the ice monitoring locations based on their seasonal ice distribution patterns. Then, we conduct experiments by training models on specific season and locations and testing them on different ones to evaluate how well the model generalizes across both spatial and temporal variations. To ensure robust validation, we reserved 10\% of the training data as a validation set. During training, we select the best-performing model based on the lowest validation loss and then evaluate its performance on test sets drawn from different locations and seasons. 

\paragraph{Structuring Seasonal Definitions for Ice Analysis} To capture the complex dynamics of ice conditions, we established two complementary seasonal classification definitions. Our first approach follows the conventional seasonal divisions—spring, summer, fall, and winter. While this categorization provides a familiar structure, it does not fully reflect the key transitions in ice-covered regions, where freeze and melt cycles drive environmental changes more significantly than calendar-based seasons. Recognizing this limitation, we developed a cryospheric seasonal classification, which directly captures the critical phases of ice formation and melting across different regions.

The cryospheric classification system is more specialized approach and introduces cryospheric seasons, which divides the year into melt and freeze periods based on location's temperature conditions. This cryospheric classification draws from the Arctic Sea Ice Melt dataset, which provides crucial information about thermal transitions in sea ice conditions. These periods are defined using the melt and freeze variables from the Arctic Sea Ice Melt dataset, which track thermal transitions. The melt season captures the period when sea ice shifts from frozen to melting, while the freeze season marks the return from melting back to frozen conditions. This dataset, part of the NASA Earth Science data collection, consists of daily averaged brightness temperature observations from the Scanning Multichannel Microwave Radiometer (SMMR) and the Special Sensor Microwave/Imager (SSM/I) sensors. The data is mapped onto a 25 km polar stereographic grid, providing high-resolution insights into sea ice thermal changes. Additionally, the dataset includes yearly maps of key ice phases: early melt (initial signs of melting), melt (sustained melting until freeze begins), early freeze (first observed freezing conditions), and freeze (continuous freezing conditions) \cite{Markus2009}.

We used Arctic Sea Ice Melt data with our existing dataset by mapping where our data overlaps with the Arctic dataset and analyzing how they relate spatially. A conservative definition of the melt and freeze seasons is to use the dataset’s melt and freeze variables, where the melt season spans from melt onset to freeze, and the ice growth season extends from freeze to the following year's melt. To determine whether a given day falls in the melt or freeze season, we compare the day of year from our scene files to the Arctic dataset's average melt and freeze dates. If the day occurs between the average melt date and average freeze date, it's classified as melt season (264 files). If the day either occurs after the average freeze date or before the average melt date, it's classified as freeze season (212 files). This classification system lets us precisely identify seasonal ice conditions for any given date in our dataset.
These definitions allow us to assess how well the model performs under varied environmental conditions, enhancing our understanding of its adaptability and effectiveness in different seasonal contexts.
\begin{figure}[!htbp]
    \centering
    \captionsetup{justification=centering} 
    \begin{minipage}{\textwidth}
        \subfloat[]{
            \includegraphics[width=0.48\textwidth]{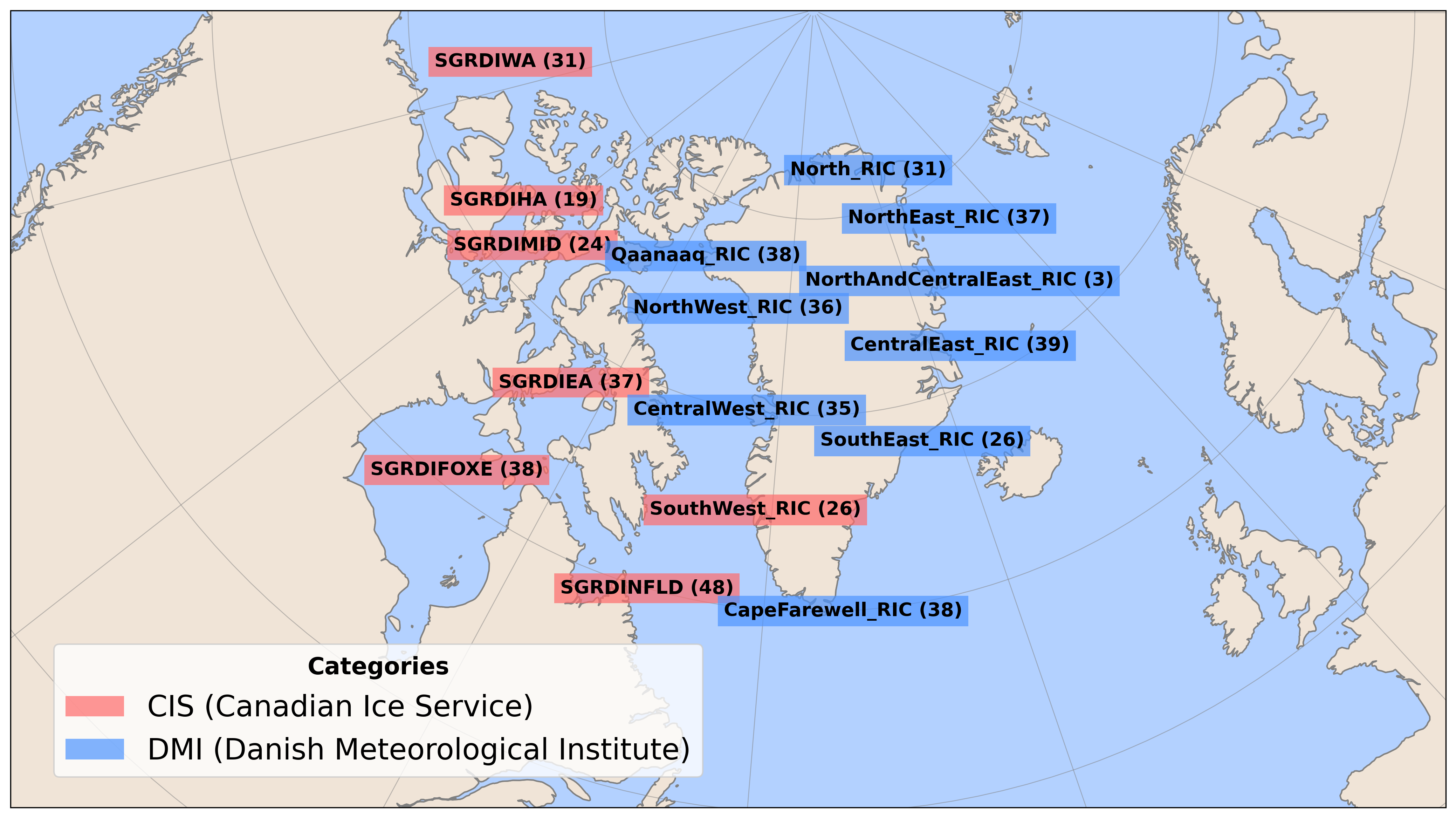}
            \label{fig:locations_png}
        }
        \hfill
        \subfloat[]{
            \includegraphics[width=0.48\textwidth]{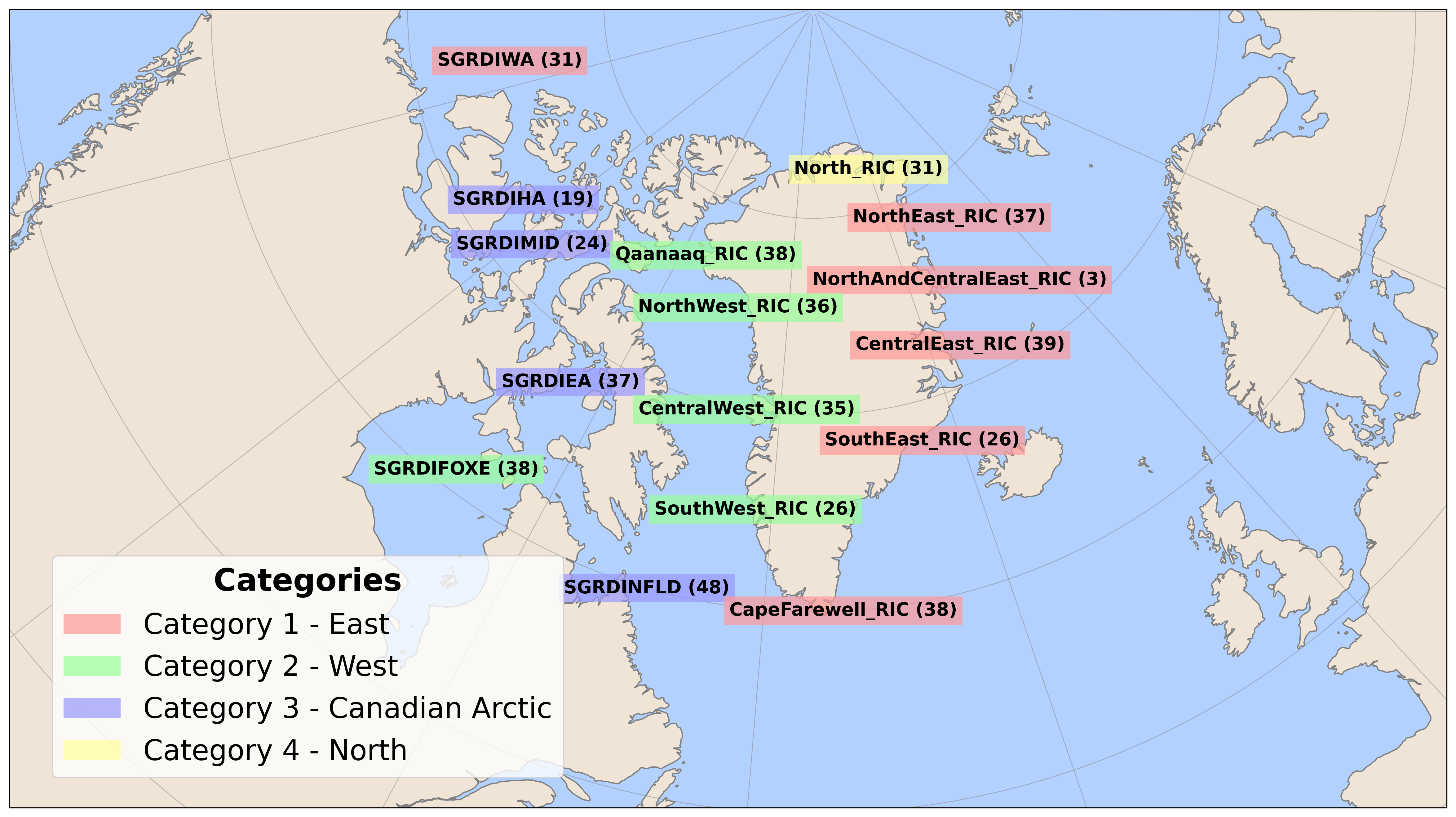}
            \label{fig:Categories_location}
        }
        \caption{(\textbf{a}) Training file distribution across CIS and DMI ice charting regions. The numbers in parentheses indicate the number of training files available for each region \cite{ChallengeDataset}.(\textbf{b}) categorization of ice regions based on seasonal ice distributions.}
        \label{fig:ice_regions}
    \end{minipage}
\end{figure}

\paragraph{Grouping Ice Regions by Seasonal Behavior} To better understand the model’s generalizability across different locations, we conducted a systematic analysis of regional variations in ice distribution. The dataset includes 16 distinct locations monitored by the CIS and DMI centers from January 2018 to December 2021, as shown in Figure \ref{fig:locations_png}. 
\begin{figure}[t]
    \centering
    \captionsetup{justification=centering} 
    \includegraphics[width = \linewidth]{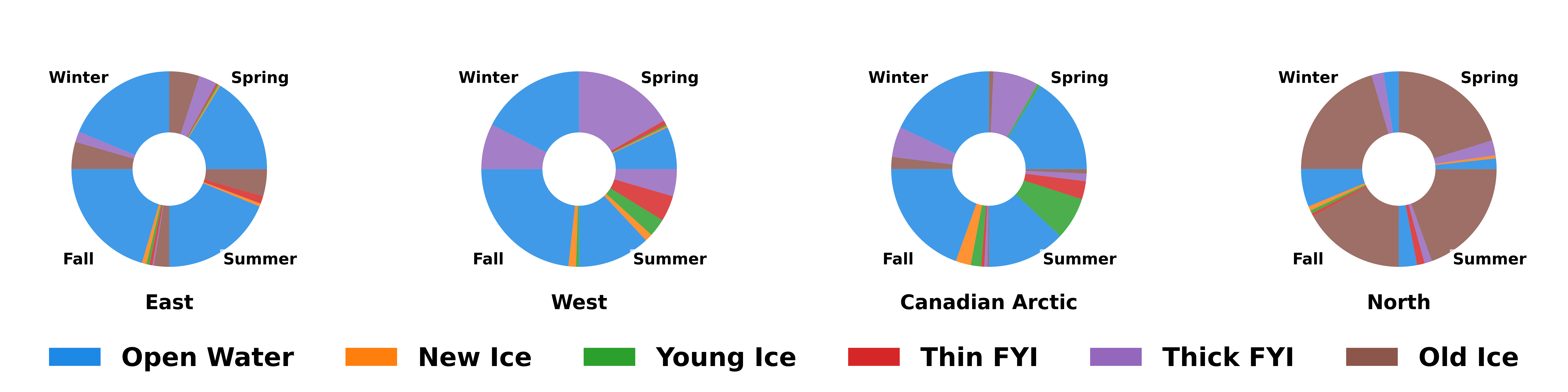} 
    \caption{Seasonal Ice Class Distributions Across Categorized Ice Regions Based on Conventional Seasons}
    \label{categories_distributions}
\end{figure}
By carefully analyzing how ice classes are distributed across different locations throughout the conventional seasons, we uncovered distinct regional ice patterns. Our approach began by quantifying the percentage of each ice class at every location across all four seasons. This analysis revealed natural groupings among locations based on their seasonal ice behavior, allowing us to classify them into four distinct categories. Figure \ref{fig:Categories_location} illustrates this classification, highlighting how ice regions were grouped based on their seasonal ice distribution patterns. To further examine these groups, Figure \ref{categories_distributions} provides a detailed visualization of the seasonal ice type distribution within each category, offering a comprehensive comparison of ice characteristics across different regions.

These categories, primarily defined by their geographic locations, exhibit markedly different seasonal patterns in ice evolution. The Eastern region (Category 1) demonstrates persistent Open water and Old ice throughout the year, with notable seasonal fluctuations in Thick First-Year Ice (FYI) while maintaining relatively stable ice class proportions. The Western region (Category 2) shows clearer seasonal transitions, dominated by Thick FYI during spring, shifting to predominantly Open water in summer and fall, with diverse ice distribution in winter months. Moving to the Canadian region (Category 3), we observe a consistent Open water presence year-round, complemented by seasonal variations where Thick FYI becomes predominant in spring and summer, Young ice prevails in winter, and New Ice forms during fall. The Northern region (Category 4) is distinguished by its substantial Old ice presence, with seasonal shifts showing increased Open water during fall and winter, and higher concentrations of Thick FYI in spring and summer.

These distinct seasonal distribution patterns across categories provide crucial insights into the regional characteristics of sea ice behavior and evolution throughout the year.


\begin{figure}[!t]
    \centering
    \includegraphics[width=\linewidth]{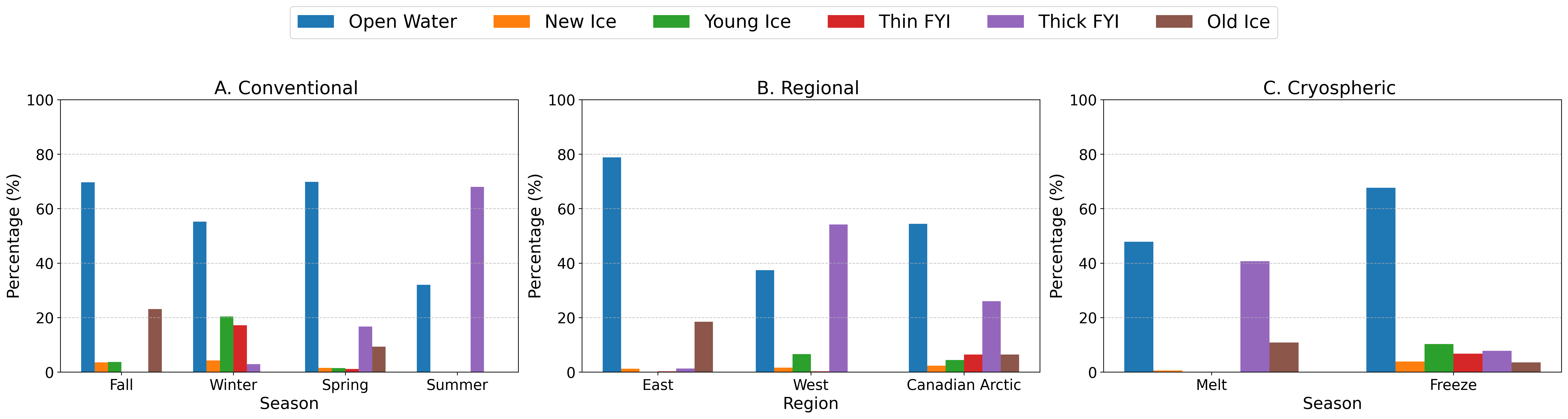} 
    \caption{Class-wise Pixel Ratio Distribution in Test Data:  
    (A) Across Conventional Seasons,  
    (B) Across Cryospheric Seasons,  
    (C) Across Geographic Regions.}
    \label{Class_wise_Pixel_Ratio_Per_Season_Test}
\end{figure}

\subsubsection{Transferability Across Conventional Seasons and Geographic Regions }
To evaluate the model’s ability to generalize across different conventional seasons and geographic regions, we designed two sets of experiments. The first experiment focuses on seasonal transferability, where the model is trained on data from a specific conventional season and then tested on different conventional seasons within the test dataset. The second experiment examines geographic transferability, where the model is trained on data from a specific geographic location and tested on different categorized locations in the test dataset. These categories correspond to the four regional groupings we previously defined.

\begin{figure}[!tb]
    \centering
    \captionsetup{justification=centering} 
    \begin{minipage}{\textwidth}
        \subfloat[]{
            \includegraphics[width=0.48\textwidth]{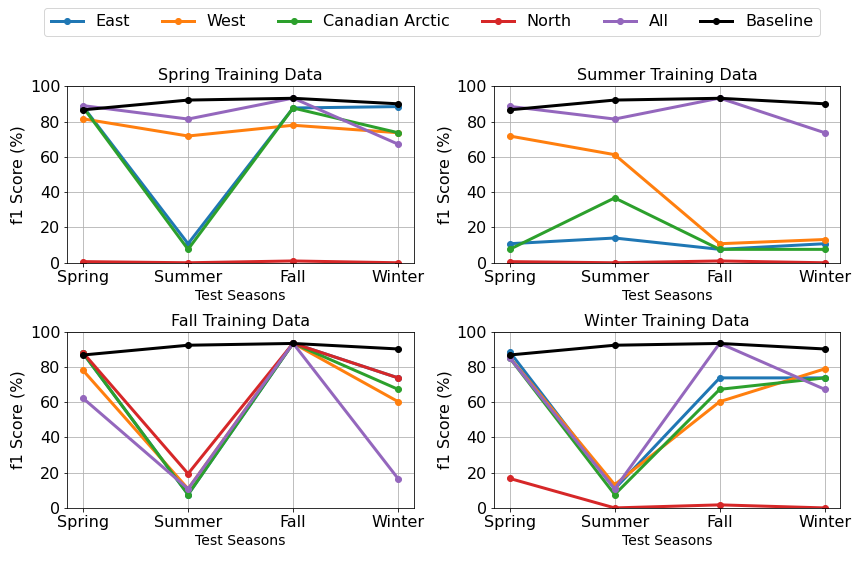}
            \label{fig:season_over_seasons_clf}
        }
        \hfill
        \subfloat[]{
            \includegraphics[width=0.48\textwidth]{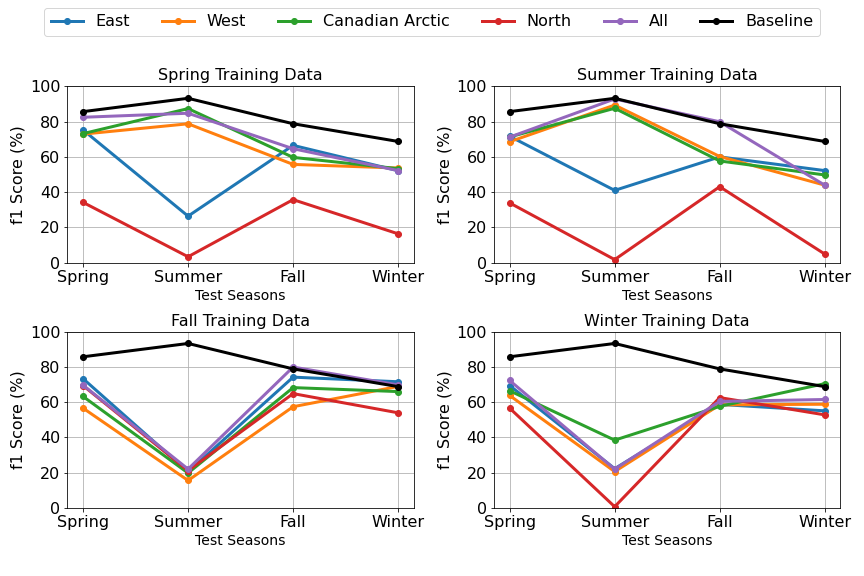}
            \label{fig:season_over_seasons_seg}
        }
        \caption{ F1-score (\%) of the model trained in a specific season (subplot title) and tested across all seasons (\textbf{a}) Patch-based model. (\textbf{b}) Pixel-based model.  
        Each line represents a training region, with the black line as the baseline.}
        \label{fig:season_comparison}
    \end{minipage}
\end{figure}

We began our transferability analysis by investigating how both patch-based and pixel-based models perform when trained on a specific conventional season and tested across conventional seasons, while also accounting for geographic variations. Figures \ref{fig:season_over_seasons_clf} and \ref{fig:season_over_seasons_seg} provide a comprehensive visualization of this interaction between seasonal and geographic factors for the Densenet and U-Net models, respectively.  These plots show how well each season‐specific model generalizes to the other seasons, providing an overall view of how sensitive the system is to shifts in seasonal conditions. Each subplot in Figures represents a model trained on a specific conventional season, as indicated in the title of the plot. The x-axis represents the seasonal data in the test set, showing how the model performs when tested on different conventional seasons. The y-axis indicates the F1-score (\%), measuring the model’s classification performance. Each colored line represents one of the four regional groups (East, West, Canadian Arctic, and North), illustrating how the model trained on a specific season performs when applied to different geographic locations. The "All" category includes all regions, meaning the model was trained on a particular season regardless of location. The "Baseline" category, shown in the legend, represents a model trained on all locations and all seasons, serving as a reference for overall performance comparison.

Model performance heavily depends on the seasonal class distribution between training and testing data. Figures \ref{categories_distributions} and \ref{Class_wise_Pixel_Ratio_Per_Season_Test}(A) show how differences in ice class distributions impact generalization. Both models perform best when tested on the same season they were trained on, but experience sharp performance drops in off-season scenarios. Models achieve higher F1 scores when tested on the same season they were trained on but experience sharp drops in off-season scenarios. These performance declines result from training-test distribution mismatches. Models adapt to the dominant ice conditions of their training season, such as thickness, open water prevalence, and regional patterns. However, testing on a different season—with more open water in summer or thicker ice in winter—introduces unseen conditions, reducing F1 scores. The model particularly struggles with underrepresented features like melt ponds and freeze-up phases, further limiting its ability to generalize. The baseline model, trained on data from all seasons and locations, achieves more consistent performance across different scenarios. This underscores the importance of diverse training data in ensuring robust model generalization.

Winter and summer test data pose challenges due to their ice compositions. Winter has more young and new ice, while summer contains more thick FYI, both of which are harder to classify. Models trained in other seasons struggle due to limited exposure to these ice types. Regional variations further impact model generalization:
Canadian Arctic and West show stable results in spring and summer, benefiting from a more balanced mix of ice types. North has high variability but performs well when trained on fall, as its train data contains more open water. It also does well on spring test data, which has thick FYI, aligning with its training conditions.
East is highly sensitive to seasonal changes, performing poorly on summer tests due to a lack of thick FYI in its training data. The models trained on different regions in fall and winter perform the worst on summer test data because the training lacks enough thick FYI. However, Canadian Arctic model trained in winter perform slightly better in summer due to some exposure to thick FYI during training.
 
 
\begin{figure}[!t]
    \centering
    \includegraphics[width=0.8\linewidth]{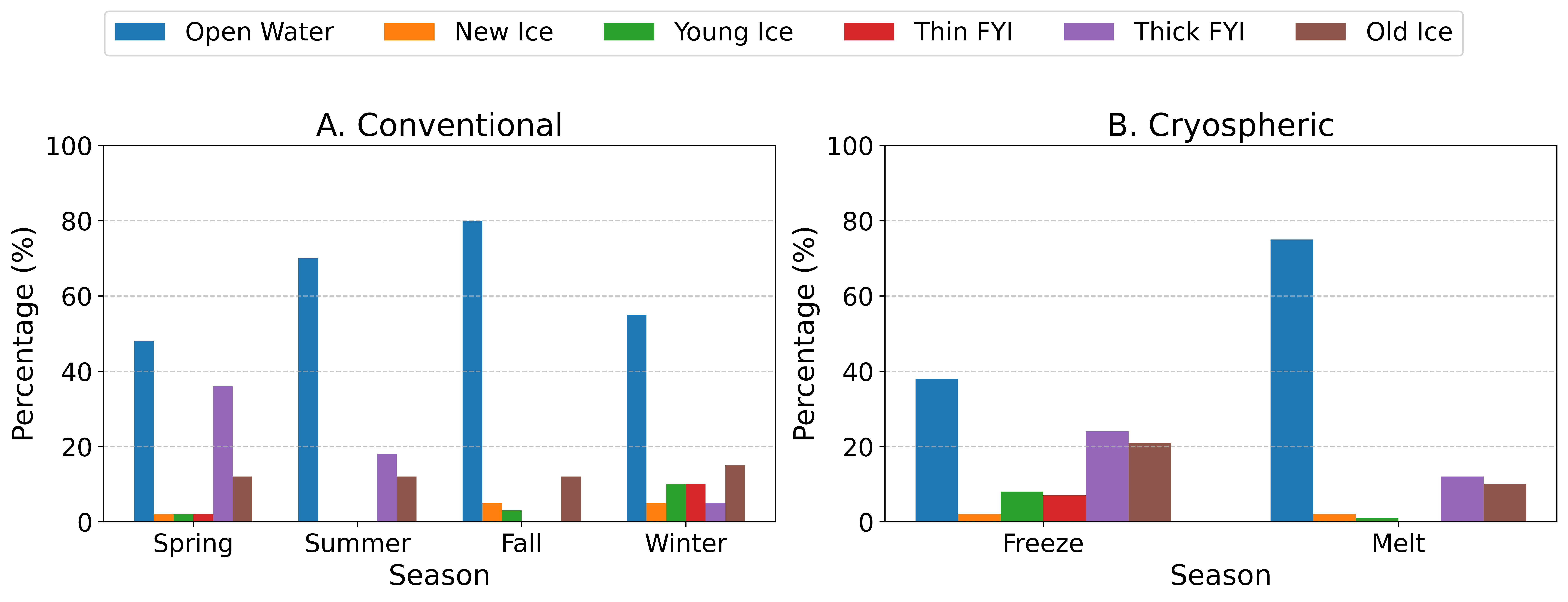} 
    \caption{Class-wise Pixel Ratio Distribution in Training Data:  
    (A) Across Conventional Seasons,  
    (B) Across Cryospheric Seasons.}
    \label{Class_wise_Pixel_Ratio_Per_Season_Train}
\end{figure}
In this analysis, we focus on the purple line ("All") in each subplot, which represents a model trained using data from all categorized locations while being specific to the season indicated in the subplot title. This allows us to evaluate how well a model trained on a particular conventional season generalizes when tested across different seasons, providing insight into the temporal transferability of the model. The U-Net and Densenet models perform best when trained on summer, followed by spring, while winter training results in the lowest performance. Figures \ref{Class_wise_Pixel_Ratio_Per_Season_Train} (A) and \ref{Class_wise_Pixel_Ratio_Per_Season_Test} (A) illustrate the seasonal distribution of ice classes, highlighting how the diversity of ice conditions in training impacts model generalization. These distributions reveal that spring and summer share similar class distributions, dominated by open water, thick FYI, and old ice, explaining their similar performance levels.

Spring-trained model: Performs well on spring and summer since the training data contains a large proportion of thick FYI and open water, which are also dominant in spring and summer test data. However, it struggles on winter due to the presence of young ice, which is less represented in the training set. Summer-trained model: Achieves its best performance on summer, followed by spring, due to the high proportion of thick FYI and open water in both seasons. Performance drops in winter, as it lacks sufficient exposure to young ice during training. Fall-trained model: Performs best on fall and spring, as the class distribution in these test seasons closely aligns with the fall training data. However, it struggles in winter, where young ice is dominant, and in summer, which has a large portion of thick FYI, making generalization difficult. Winter-trained model: Surprisingly, achieves its highest performance on spring, likely due to similar class distributions between winter training data and spring in test which are thick FYI and old ice. Performance on fall is also reasonable, but it struggles with summer, which contains a significant proportion of thick FYI, a condition it has not encountered as frequently in training.

In general, models trained in spring and summer demonstrate moderate generalizability and perform well in seasons with similar class distributions.

\begin{figure}[!tb]
    \centering
    \captionsetup{justification=centering} 
    \begin{minipage}{\textwidth}
        \subfloat[]{
            \includegraphics[width=0.48\textwidth]{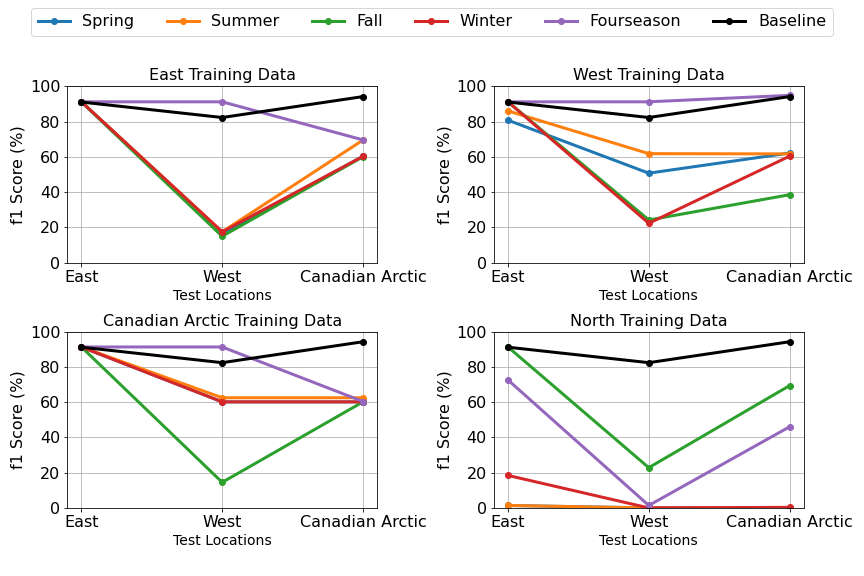}
            \label{fig:locations_over_locations_clf}
        }
        \hfill
        \subfloat[]{
            \includegraphics[width=0.48\textwidth]{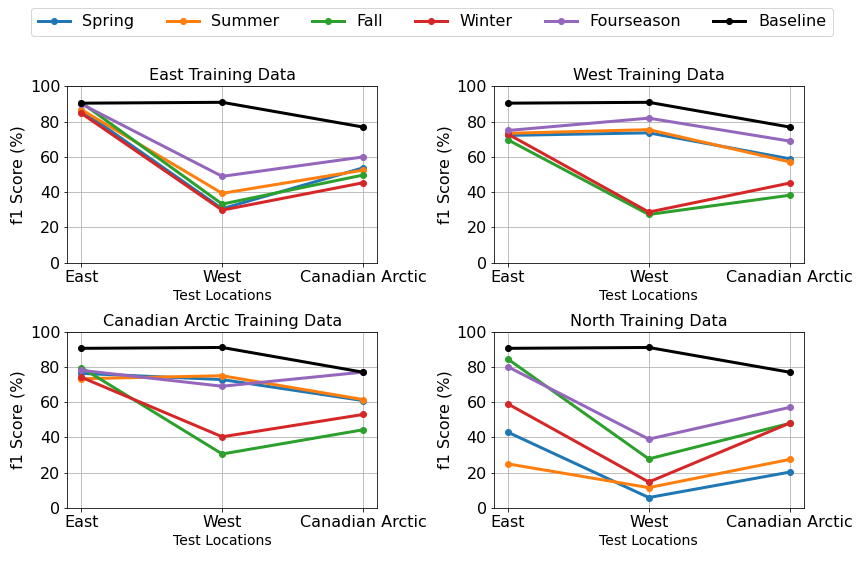}
            \label{fig:locations_over_locations_seg}
        }
        \caption{F1-score (\%) of the model trained on a specific geographic location (subplot title) and tested across all locations. (\textbf{a}) Patch-based model. (\textbf{b}) Pixel-based model. Each line represents a training season, with the black line as the baseline.}
        \label{fig:locations_comparison}
    \end{minipage}
\end{figure}

After analyzing seasonal transferability, we examined geographic transferability by training models on specific categorized locations and evaluating their performance across different regional groups. Figures \ref{fig:locations_over_locations_clf} and \ref{fig:locations_over_locations_seg} visualize this analysis for the Densenet and U-Net models, respectively. Each subplot represents a model trained on a specific categorized location, as indicated in the title of the plot. The x-axis denotes the test locations, showing how well the model performs when tested on different regional groups. The y-axis represents the F1-score (\%), measuring classification performance. Each colored line corresponds to a specific conventional season, as shown in the legend. This allows us to observe how well models trained on certain geographic regions adapt to seasonal variations in different locations. The "Fourseason" category represents a model trained on all four conventional seasons combined, rather than a single season, while the "Baseline" category refers to a model trained on all locations and all seasons, serving as a general reference for performance comparison.It is important to note that the test dataset does not contain any files from the North (Category 4) locations, and therefore, this category is not represented in the results. Figures \ref{categories_distributions} and \ref{Class_wise_Pixel_Ratio_Per_Season_Test}(C) show how differences in ice class distributions impact generalization.

The patch-based Densenet model, as shown in Figure \ref{fig:locations_over_locations_clf}, exhibits strong regional specificity in its performance. Models trained on data from specific regions achieve optimal performance when tested on data from the same region but show noticeable degradation when applied to different regions. The East test location emerges as a particularly interesting case, demonstrating consistently robust performance across different training scenarios, suggesting regional characteristics that facilitate better model generalization. However, when tested on West and Canadian Arctic locations, performance declines notably, especially during the Summer season. The West test location shows strongest performance with West-specific training data, highlighting effective within-region generalization, while showing diminished performance for East and Canadian Arctic test locations.
The pixel-based U-Net model demonstrates different geographic adaptation patterns, as illustrated in Figure \ref{fig:locations_over_locations_seg}. The East test region maintains superior performance regardless of training location, indicating robust feature characteristics in this region. West-trained models achieve optimal results within their home region, matching East region performance levels, but struggle significantly with Canadian Arctic data. Notably, this strong home-region performance shows seasonal variation, with deterioration during Fall and Winter seasons. Interestingly, models trained on Canadian Arctic data show unexpected excellence in East region predictions, while North-trained models perform best in East regions and demonstrate moderate success in Canadian regions compared to their West region performance.

\subsubsection{ Transferability Across Cryospheric Seasons and Geographic Regions} 
Expanding our analysis from conventional seasons to cryospheric seasons, we conducted the same set of experiments as defined for conventional season transferability, but instead using melt and freeze periods. These experiments reveal distinct patterns in both Densenet and U-Net models, providing insight into their ability to generalize across seasonal transitions and geographic regions in the context of cryospheric seasons.

\begin{figure}[!tb]
    \centering
    \captionsetup{justification=centering} 
    \begin{minipage}{\textwidth}
        \subfloat[]{
            \includegraphics[width=0.48\textwidth]{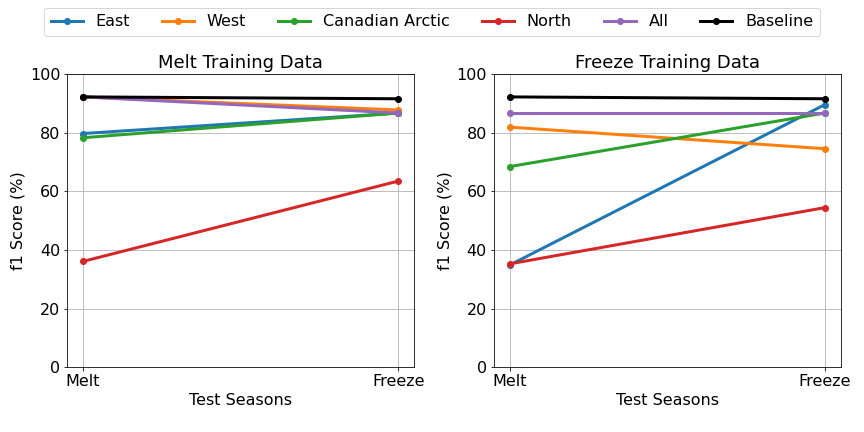}
            \label{fig:twoseason_over_twoseasons_clf}
        }
        \hfill
        \subfloat[]{
            \includegraphics[width=0.48\textwidth]{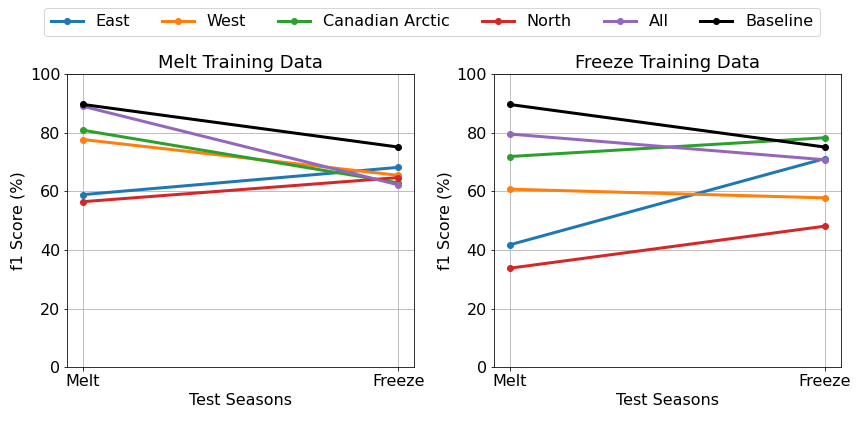}
            \label{fig:twoseason_over_twoseasons_seg}
        }
        \caption{F1-score (\%) of the model trained on a cryospheric season (subplot title) and tested across cryospheric seasons. (\textbf{a}) Patch-based model. (\textbf{b}) Pixel-based model. Each line represents a different training region.}
        \label{fig:twoseason_comparison}
    \end{minipage}
\end{figure}

For the seasonal transferability experiment using cryospheric seasons, we evaluated how well models trained on one cryospheric season (melt or freeze) perform when tested on the other. Figure \ref{fig:twoseason_over_twoseasons_clf} presents the Densenet model’s performance, while Figure \ref{fig:twoseason_over_twoseasons_seg} shows corresponding results for the U-Net model. Each subplot represents a model trained on a melt or freeze cryospheric season, while the x-axis denotes the test season, showing how the model performs when tested on a different cryospheric season. The y-axis indicates the F1-score (\%), measuring classification performance. Each colored line corresponds to a specific categorized location, illustrating how geographic variations influence the model’s ability to transfer knowledge across cryospheric seasons. Additionally, two reference categories are included: "All" (purple) represents a model trained on data from all categorized locations within a specific cryospheric season, and "Baseline" (black) represents a model trained on all locations and both cryospheric seasons, serving as a general performance reference.

The patch-based model demonstrates varying transferability patterns across cryospheric seasons. When trained on melt data, the model maintains robust F1-scores (79-90\%) for East, West, and Canadian Arctic regions across both test seasons. However, the aggregate all category shows a striking contrast, dropping from 80\% during melt testing to 45\% during freeze testing, suggesting potential overfitting to melt-specific features. Interestingly, the North region exhibits an unexpected improvement from 30\% during melt testing to 65\% during freeze testing. This enhanced performance during freeze testing can be attributed to the higher proportion of open water in freeze test files, which typically presents a simpler classification task. Conversely, the more complex mix of open water and thick FYI in melt test data creates a more challenging classification scenario.
Models trained on freeze data show different adaptation patterns. The East, Canadian Arctic, and North regions demonstrate significant improvement from melt to freeze testing periods. The West region maintains relatively stable performance across both test seasons. Notably, the baseline performance remains consistently high across both training scenarios, indicating robust overall generalization. The pixel-based U-Net model exhibits distinct regional patterns compared to the Densenet approach. Under melt data training, the West, Canadian Arctic, and aggregate regions show a gradual decline in performance from melt to freeze test seasons, while the East and North regions maintain more stable performance with slight improvements.

In this analysis, we conduct a temporal evaluation by examining the purple line ("All") in each subplot, which represents a model trained on data from all categorized locations while being specific to the cryospheric season indicated in the subplot title. This allows us to assess the model’s ability to generalize across melt and freeze periods, providing insights into its seasonal transferability. To better understand how seasonal variations impact model performance, we analyze the distribution of ice classes across melt and freeze periods, as shown in Figures \ref{Class_wise_Pixel_Ratio_Per_Season_Train} (B) for the train and Figure \ref{Class_wise_Pixel_Ratio_Per_Season_Test} (B) for test. These class distributions highlight key differences in ice conditions between seasons, which directly affect model adaptability. The pixel-based classification model demonstrates strong seasonal stability, maintaining consistent performance across all evaluation metrics in both melt and freeze periods. This stability is further enhanced when the model is trained on data that combines both seasons, suggesting that pixel-level features remain relatively stable across cryospheric transitions. In contrast, the patch-based classification model exhibits greater seasonal sensitivity. It achieves optimal performance ( $\sim 90\%$ across metrics) when trained on melt data or a combined seasonal dataset. However, when trained solely on freeze season data, performance declines significantly to 65-70\%. Furthermore, as shown in Figure \ref{Class_wise_Pixel_Ratio_Per_Season_Test} (B), the classes of new ice and young ice are significantly less present in the melt test data, making it easier for the model to achieve higher performance on melt season predictions. These ice types are particularly challenging for the model due to their variability and transitional nature, which makes their lower prevalence in the melt season beneficial for model accuracy. The increased presence of new and young ice in the freeze test data likely contributes to the performance drop when the model is trained exclusively on freeze season data, as these classes introduce more complexity and uncertainty in the classification process.

\begin{figure}[!t]
    \centering
    \captionsetup{justification=centering} 

    \begin{minipage}{0.48\textwidth} 
        \centering
        \subfloat[]{
            \includegraphics[width=\linewidth]{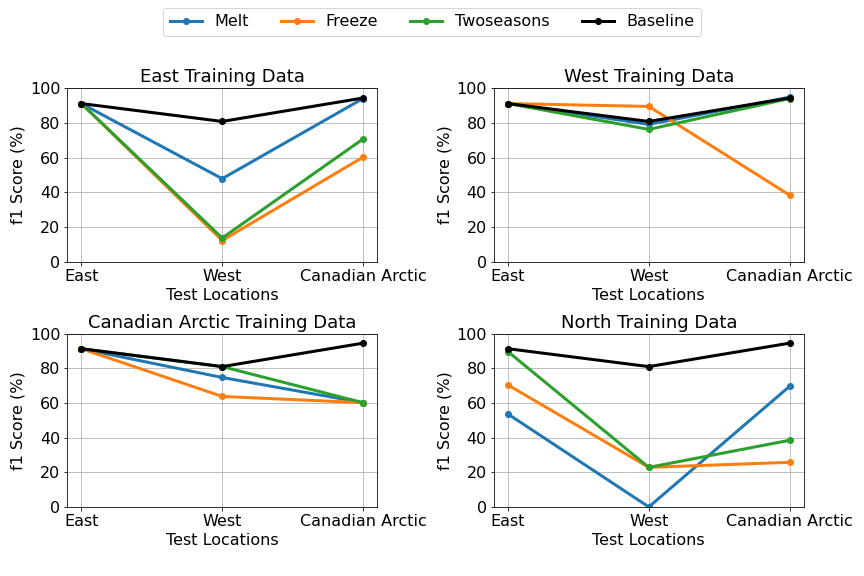}
            \label{fig:location_over_location_twoseason_clf}
        }
    \end{minipage}
    \hfill
    \begin{minipage}{0.48\textwidth} 
        \centering
        \subfloat[]{
            \includegraphics[width=\linewidth]{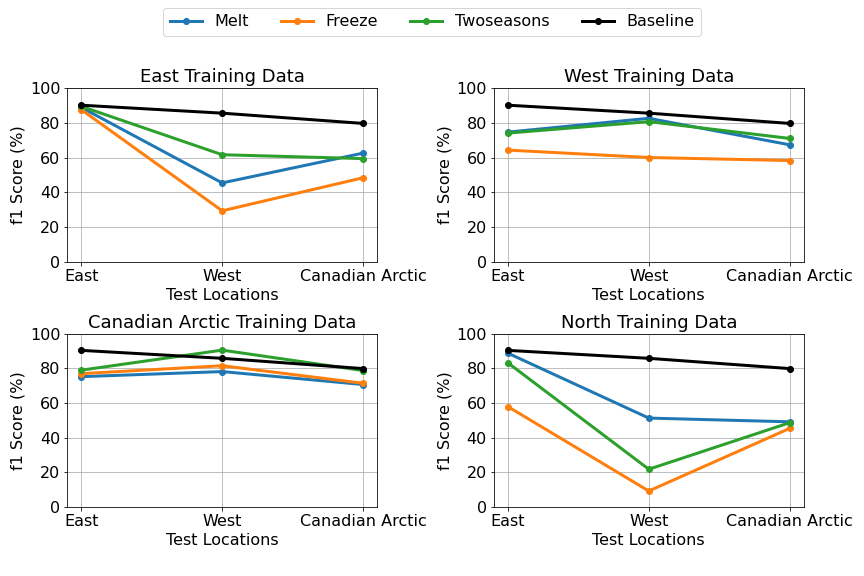}
            \label{fig:location_over_location_twoseason_seg}
        }
    \end{minipage}

    \caption{F1-score (\%) comparison of models trained on different locations (subplot title) and tested across different locations.  
    (\textbf{a}) Patch-based model. (\textbf{b}) Pixel-based model. Each line represents a different training cryospheric season.}
    
    \label{fig:location_twoseason_comparison}
\end{figure}

Our analysis of geographic transferability under cryospheric seasonal conditions examines how models trained on specific categorized locations perform when tested on different regional groups during melt and freeze periods. Figures \ref{fig:location_over_location_twoseason_clf} and \ref{fig:location_over_location_twoseason_seg} illustrate these results for the Densenet and U-Net models, respectively. Each subplot represents a model trained on a specific categorized location, while the x-axis indicates the test location, showing performance across different regional groups. Each colored line corresponds to a specific cryospheric season melt and freeze demonstrating how the model trained on one location performs when tested on different regions during each season. The "Twoseasons" category refers to a model trained on both melt and freeze seasons for a specific location. Figures \ref{categories_distributions} and \ref{Class_wise_Pixel_Ratio_Per_Season_Test}(C) help illustrate regional ice class distributions.
The pixel-based model shows distinct regional adaptation patterns. The East-trained model achieves impressive F1-scores of approximately 90\% on its home region but experiences significant performance degradation (40-60\%) when tested on West and Canadian Arctic regions. In contrast, the West-trained model emerges as the most robust, maintaining consistent F1-scores between 60-80\% across all regions, demonstrating strong generalization capabilities. The Canadian Arctic-trained model shows interesting behavior, performing exceptionally well on both East and its home region (80-90\% F1-scores) and achieving its best performance on the West region. The North-trained model exhibits high variability, excelling in the East region (80-90\% F1-scores) but struggling with Canadian Arctic and West regions (20-50\%).

For patch-based classification, we observe different regional adaptation characteristics. The East-trained model demonstrates near-perfect performance in its home region but struggles significantly with the West region, while showing improved performance (60-90\% F1-scores) on Canadian Arctic data. The West-trained model shows remarkable generalization, excelling not only in its home region but also in East and Canadian Arctic regions, particularly during the melt season. The Canadian Arctic-trained model maintains consistently high performance (60-90\% F1-scores) on both East and West regions but, surprisingly, shows lower performance in its home region. The North-trained model achieves high performance (80-100\% F1-scores) on East region data but demonstrates poor generalization to West and Canadian Arctic regions (20-40\%).

Across both approaches, West region emerge as the most generalizable, suggesting that it captures a diverse range of characteristics of sea ice applicable across different regions. East and North regions show similar patterns in both tasks, with models trained on these regions generalizing poorly to other areas but showing some mutual compatibility. Canadian Arctic-trained models often perform well on regions other than their training area, suggesting they may capture intermediate or shared features across Arctic regions. These findings highlight the importance of considering regional characteristics in model development and deployment strategies for Arctic sea ice classification.

\subsection{Exploring the Impact of Downscaling the Resolution }
The downscaling ratio of images significantly influences model performance by balancing computational efficiency with detail preservation. While high-resolution scenes contain more detailed information critical for accurate classification in sea ice monitoring, they demand greater computational resources. 

\begin{figure}[!tb]
    \centering
    \captionsetup{justification=centering}  
    \begin{minipage}{\textwidth}
        \subfloat[]{
            \includegraphics[width=0.48\textwidth]{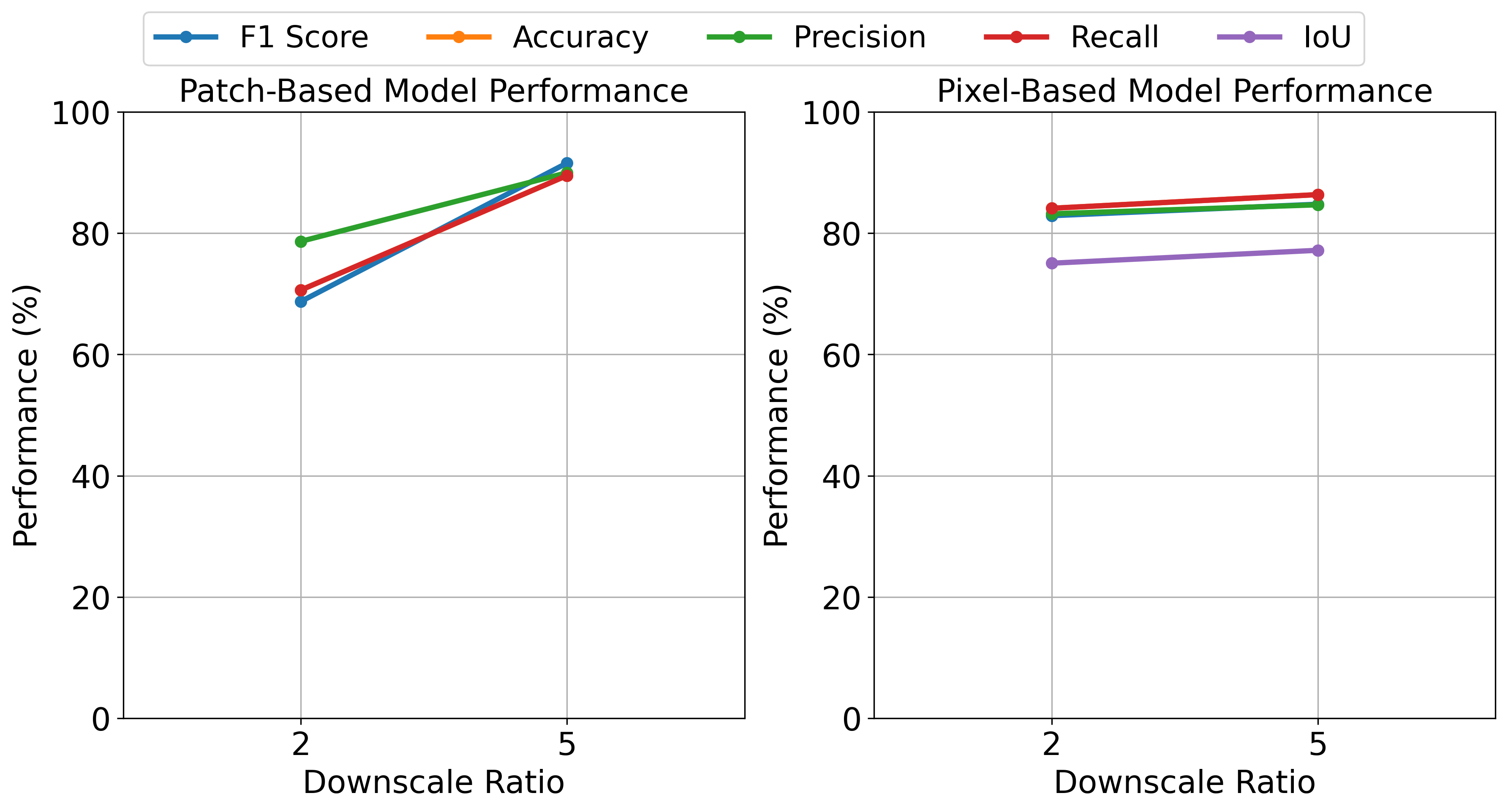}
            \label{fig:downscale_ratio}
        }
        \hfill
        \subfloat[]{
            \includegraphics[width=0.48\textwidth]{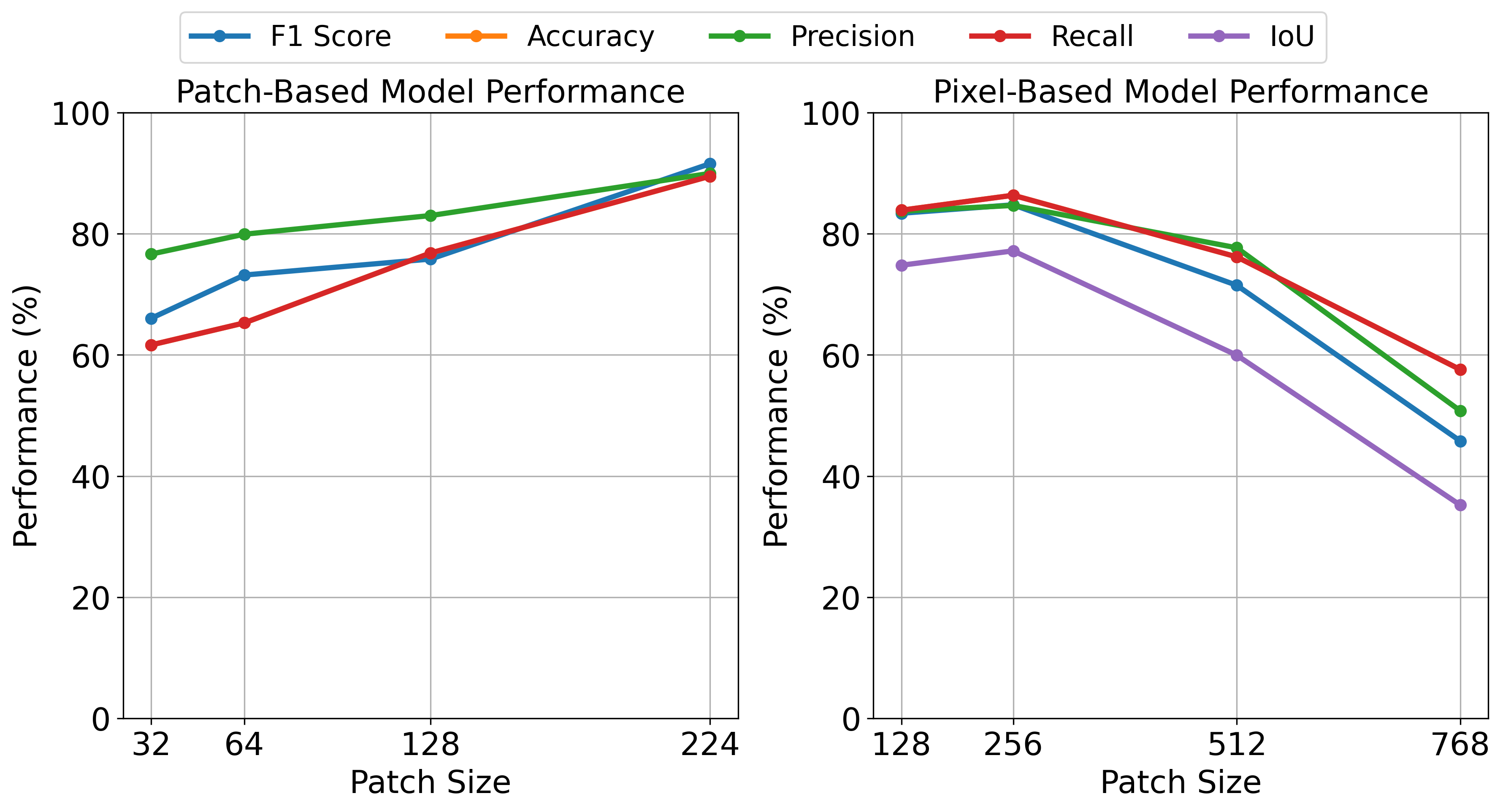}
            \label{fig:patch_size}
        }
        \caption{Performance comparison of patch-based and pixel-based models across different configurations:  
        (\textbf{a}) Downscaling ratios and (\textbf{b}) Patch sizes.}
        \label{fig:downscale_patch_comparison}
    \end{minipage}
\end{figure}

Figure \ref{fig:downscale_ratio} presents two subplots comparing pixel-based and patch-based model performance across different downscaling ratios. The x-axis represents the tested ratios (2 and 5), while the y-axis shows accuracy metrics (\%), with each line indicating a different accuracy metric.

Our experimental results demonstrate that for patch-based classification, a downscaling ratio of 5 yielded superior results compared to a ratio of 2. This suggests that moderate resolution reduction does not significantly compromise the model's ability to identify larger-scale patterns in the sea ice imagery. In pixel-based classification tasks, both downscaling ratios (2 and 5) demonstrated comparable performance, indicating that pixel-level features remain relatively preserved even at lower resolutions.
This performance pattern can be interpreted through the lens of feature preservation versus computational efficiency. While higher resolutions theoretically retain more detailed information, our results suggest that a moderate reduction in resolution can maintain classification accuracy while significantly reducing computational overhead. This finding has practical implications for deploying these models in resource-constrained environments or real-time applications.

A notable observation in our study concerns the case of downscaling ratio 0 (original resolution). Despite the theoretical advantage of maximum detail preservation, we had to exclude this configuration from our analysis due to its prohibitive computational requirements. This exclusion highlights the practical limitations that must be considered when deploying deep learning models in real-world applications.

\subsection{Exploring the Impact of Patch Size }
Patch size defines the dimensions of sub-images extracted from the larger image, influencing the trade-off between fine-scale detail and broader spatial context. Smaller patches capture intricate ice features but may miss large-scale patterns, while larger patches provide more context but can overlook finer details. To ensure a fair evaluation, we selected patch sizes of 32 to 224 pixels for patch-based models and 128 to 768 pixels for pixel-based models, reflecting common practices in each approach. 

Figure \ref{fig:patch_size} presents two subplots comparing pixel-based and patch-based model performance. The x-axis represents patch size, while the y-axis shows accuracy metrics \%) with different lines indicating various metrics, as per the legend. The pixel-based classification subplot shows optimal performance at a moderate patch size of 256 pixels, with metrics declining for larger patches. As patch sizes increase from 32 to 224 pixels, all evaluation metrics show consistent improvement. This upward trend suggests that patch-based approaches benefit from larger contextual windows, which likely provide richer spatial information for classification decisions. 

Patch-based method show continuous performance improvements as patch sizes increase to 224 pixels, suggesting they effectively utilize broader contextual information. In contrast, pixel-based method perform optimally at moderate patch sizes (256 pixels) before declining with larger contexts. These divergent behaviors highlight the importance of selecting appropriate patch sizes based on the classification approach.


\subsection{Exploring the Impact of Data Size }
\begin{figure}[!tb]
    \includegraphics[width= \linewidth ]{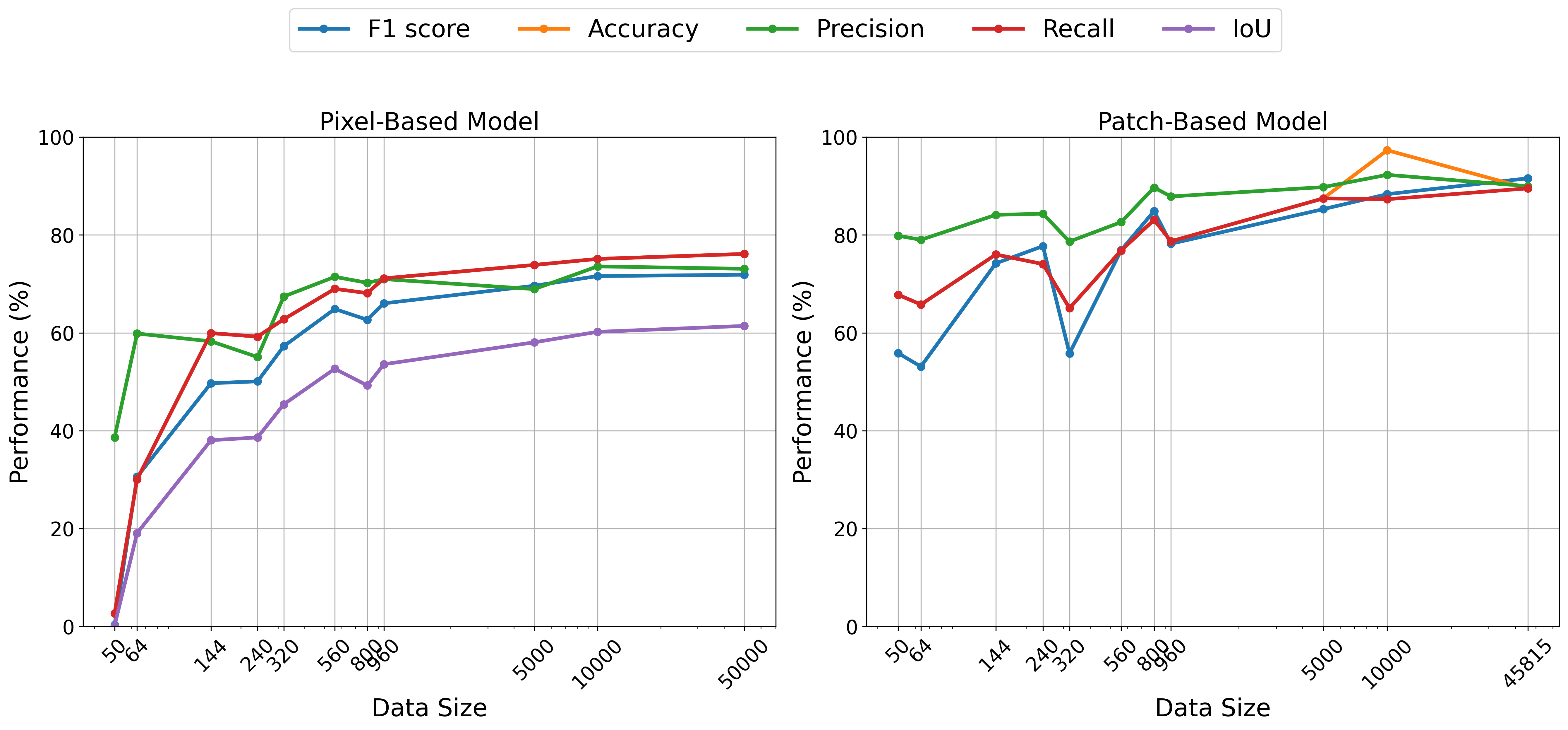} 
    \centering
    \caption{Performance of patch-based and pixel-based models across different data sizes
    }
    \label{data_size}
\end{figure}

Data size is critical for training deep learning models, particularly for sea ice classification, where diverse and representative samples enhance generalization. The patch-based model uses pre-generated patches, while for this experiment, the pixel-based model was trained on patches generated with a patch size of 256 and a stride of 100, rather than using random cropping. Figure \ref{data_size} consists of two subplots, each illustrating the effect of dataset size on the pixel-based and patch-based models. The x-axis represents the number of training samples, while the y-axis shows performance metrics (\%), with different lines corresponding to various accuracy measures.

The pixel-based model demonstrates a consistent and gradual learning trajectory. Beginning with minimal effectiveness at small data sizes (50-64 samples), it exhibits steady improvement as the training volume increases. Performance metrics generally plateau after approximately 5,000-10,000 samples, indicating the model reaches its learning capacity at this threshold, with additional data offering diminishing returns. All evaluation metrics for the pixel-based model follow similar growth patterns, although IoU consistently remains lower than other measurements throughout the training process.
In contrast, the patch-based model achieves strong performance even with limited training data and improves further with larger datasets. This early advantage suggests that the patch-based approach captures relevant features more efficiently. However, its learning curve is more variable, with fluctuations such as an F1 score dip around the 240–320 sample range, possibly due to batch composition effects.


\begin{table}[H]
\centering
\caption{Classification results of pixel-based and patch-based models with different data preparation strategies.}
\label{tab:data_prep_combined}
\resizebox{\textwidth}{!}{%
\begin{tabular}{l c c c c c | c c c c}
\toprule
\multirow{2}{*}{\textbf{Scenario}} & \multicolumn{5}{c|}{\textbf{Pixel-Based Model}} & \multicolumn{4}{c}{\textbf{Patch-Based Model}} \\
\cmidrule(lr){2-6} \cmidrule(lr){7-10}  
 & \textbf{F1-score} & \textbf{Accuracy} & \textbf{Precision} & \textbf{Recall} & \textbf{IoU}  
 & \textbf{F1-score} & \textbf{Accuracy} & \textbf{Precision} & \textbf{Recall} \\  
\midrule
\multicolumn{10}{c}{\textbf{With/Without Data Augmentation}} \\  
With Aug. & 84.78 & 86.36 & 84.68 & 86.36 & 77.18  
& 91.57 & 89.51 & 89.97 & 89.51 \\  
Without Aug. & 83.44 & 83.70 & 84.89 & 83.70 & 75.41  
& 81.92 & 76.54 & 89.38 & 76.54 \\  
\midrule
\multicolumn{10}{c}{\textbf{Removing/Including Land Pixels}} \\  
Remove Land Pixels & 83.51 & 84.35 & 83.27 & 84.35 & 74.73  
& 81.92 & 76.54 & 89.38 & 76.54 \\  
Include Land Pixels & 84.78 & 86.36 & 84.68 & 86.36 & 77.18  
& 91.57 & 89.51 & 89.97 & 89.51 \\  
\midrule
\multicolumn{10}{c}{\textbf{Distance to Border (Threshold 20)}} \\  
Distance to Border & 67.81 & 72.12 & 70.78 & 72.12 & 55.94  
& 79.27 & 79.19 & 87.77 & 79.19 \\  
\bottomrule
\end{tabular}%
}
\end{table}

\subsection{Exploring the Impact of Data Preparation Methods}
Previous studies have demonstrated that variations in preprocessing techniques can significantly influence model generalization, feature extraction, and robustness. Motivated by these findings, we systematically evaluated key data preparation strategies to assess their impact on both patch-based and pixel-based classification tasks. As shown in Table \ref{tab:data_prep_combined}, we examined three major data preparation methods: data augmentation, land pixel inclusion, and distance-to-border thresholding. Each technique addresses specific challenges in sea ice image analysis and contributes to the overall robustness of the classification models.

First and foremost, data augmentation is a widely used technique that artificially increases the diversity of the training dataset by applying transformations such as rotation, flipping, scaling, and cropping. This helps the model generalize better by exposing it to different variations of the data. The results presented in Table \ref{tab:data_prep_combined} for patch-base model demonstrate its effectiveness for patch-based model, where data augmentation substantially improved performance, elevating the F1-score from 81.92\% to 91.57\%. For pixel-based model, shown in Table \ref{tab:data_prep_combined} for pixel-based, the impact was more modest but still positive, with the F1-score increasing from 83.44\% to 84.78\%. These improvements indicate that augmentation helps models develop better generalization capabilities by exposing them to diverse ice conditions and mitigating overfitting issues.

Additionally, the consideration of distance to the border represents another technique evaluated for its impact on model performance. Since sea ice labels are defined at the polygon level in ice charts, distance to border refers to maintaining a minimum distance from the polygon boundaries when selecting training patches. This approach aims to increase patch purity by excluding ambiguous regions near class transitions.However, Table \ref{tab:data_prep_combined} show that applying a 20-pixel distance-to-border threshold had a negative impact on both classification approaches. For patch-based classification, the F1-score declined to 79.27\%, indicating a performance drop due to boundary removal. DenseNet’s feature reuse makes it powerful but sensitive to data changes, such as boundary removal. In pixel-based classification, the decrease was more significant, falling to 67.81\%. This deterioration in performance suggests that edge regions provide essential information for accurate classification, particularly in pixel-based tasks where precise boundary determination is crucial.

\begin{figure}[!t]
    \includegraphics[width=\textwidth   ]
    {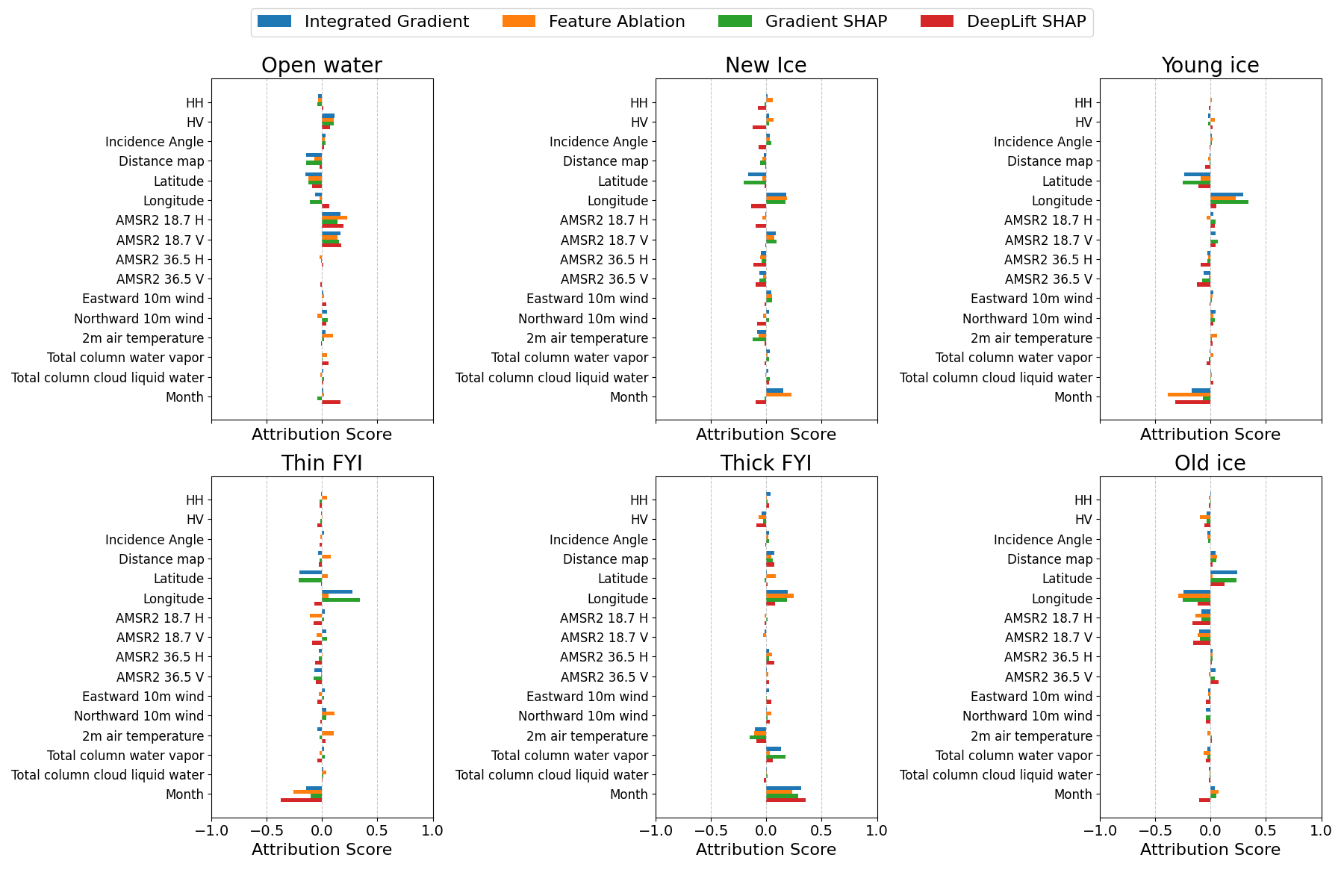} 
    \centering
    \caption{Feature importance analysis for six classes using the patch-based model}
    \label{feature_importance_clf}
\end{figure}
Furthermore, land pixel removal is another technique evaluated for its impact on model performance. By filtering out these pixels, we ensure that the model focuses solely on the relevant sea ice data. Interestingly, as evident in Tables \ref{tab:data_prep_combined}, the inclusion of land pixels emerged as a beneficial factor for both classification approaches. In patch-based classification, this inclusion led to a substantial improvement in the F1-score from 81.92\% to 91.57\%, accompanied by significant enhancements in accuracy and recall. DenseNet's dense connectivity makes it sensitive to data distribution. Removing land pixels with contextual cues can disrupt feature reuse and degrade performance. The pixel-based classification showed similar benefits, with the F1-score rising from 83.51\% to 84.78\%. These improvements suggest that land pixels provide valuable contextual information, particularly at coastline interfaces where the distinction between land, water, and ice types is critical.


\subsection{Feature Importance Analysis}
Understanding how different input features influence model behavior is crucial for both model interpretation and validation in sea ice classification. Our analysis employs multiple attribution methods to quantify and compare feature importance across both approaches, providing insights into how each model makes decisions across six different ice types.
\begin{figure}[!t]
   
    \includegraphics[width=\textwidth   ]{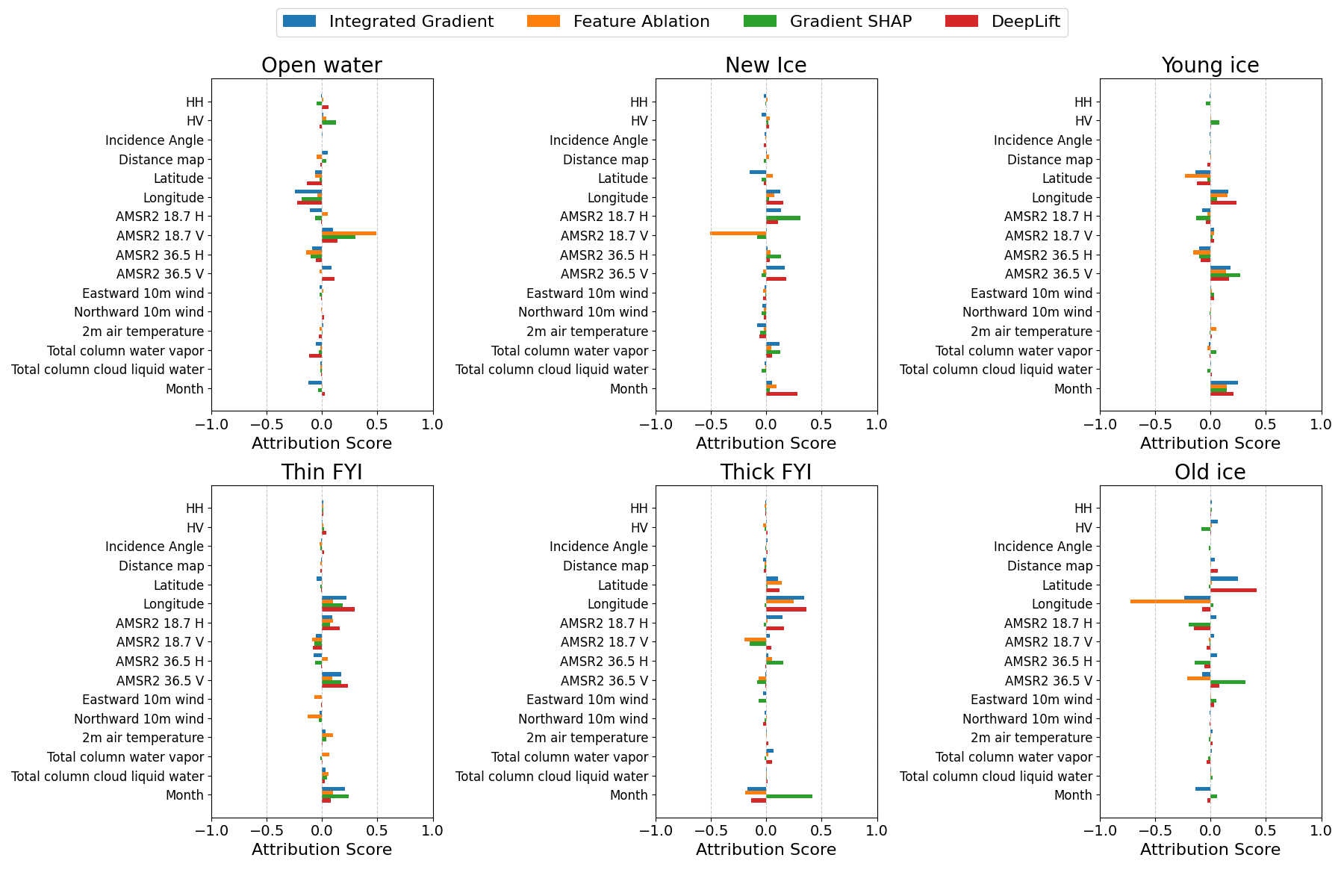} 
 
    \centering
    \caption{Feature importance analysis for six classes using the pixel-based model}
    \label{feature_importance_Seg}
\end{figure}
For this investigation, we employed Captum, a comprehensive model interpretability library. The analysis incorporates four attribution methods: Integrated Gradient, which computes attribution by integrating gradients along a specified path from baseline to input; Feature Ablation, which systematically removes features to measure their impact; Gradient SHAP, which applies Shapley values from cooperative game theory; and DeepLift SHAP, which combines Shapley values with the DeepLift framework to assess feature contributions relative to a reference point \cite{Lundberg2017}.

The patch-based classification model's feature importance analysis, shown in Figure \ref{feature_importance_clf}, reveals distinctive patterns across different ice classes. In the plot, HH and HV stand for the SAR primary (nersc\_sar\_primary) and SAR secondary (nersc\_sar\_secondary), respectively.

Longitude emerges as a dominant feature, demonstrating consistently high importance across most ice types. This geographical dependency suggests that spatial location plays a fundamental role in ice type determination. The model exhibits varying reliance on features such as distance maps and temperature metrics, indicating these parameters have specialized relevance for specific ice conditions. Seasonal patterns, captured through the "month" feature, show substantial importance in certain ice classifications, highlighting the temporal dynamics of ice formation and transformation. Temperature-related features and meteorological variables like total column water vapor and liquid water demonstrate varying significance, particularly in identifying thick FYI. Notably, some features display negative attribution scores, indicating that their absence serves as evidence against particular ice classifications, demonstrating the model's sophisticated decision-making process. 

The segmentation model's feature importance results, illustrated in Figure \ref{feature_importance_Seg}, present distinct patterns from the classification model. For open water detection, the model primarily relies on brightness temperature features, specifically AMSR2 18.7 GHz vertical and AMSR2 36.5 GHz horizontal polarizations, along with SAR-derived features. These spectral and backscatter characteristics prove crucial for distinguishing open water from ice surfaces.
Similar to the classification model, geographical features maintain high importance across ice classes in the segmentation model, reinforcing the critical role of spatial information in ice type determination. The temporal component, represented by the month feature, demonstrates significant attribution scores across multiple ice types, capturing the seasonal variations in ice dynamics. While meteorological variables such as 2-meter air temperature and atmospheric water content show relatively lower attribution scores, they contribute meaningful refinements to the segmentation process.

The comparison between patch-based and pixel-based models reveals both shared and distinct patterns in feature utilization. While both models heavily rely on geographical features, the pixel-based model shows greater sensitivity to spectral characteristics, particularly in open water detection. The patch-based model demonstrates more nuanced use of meteorological variables, while the pixel-based model places greater emphasis on direct observational data from SAR and AMSR2 sensors. These differences reflect the complementary nature of the two approaches, each optimized for their specific task in sea ice analysis.

\section{Discussion and Conclusion}

We introduced a comprehensive IceBench framework designed to evaluate the performance of deep learning models in the context of sea ice type classification. This IceBench provides a systematic approach to assess various aspects of model performance, including accuracy and efficiency. The primary objective of our IceBench is to provide clear, reproducible metrics that allow for the comparison of different models and techniques. 

The findings from our IceBench highlight several key insights into model performance and evaluation. One primary observation is the sensitivity to hyperparameters such as learning rate and batch size. Our results show that a moderate learning rate combined with a larger batch size balances learning efficiency and computational stability.
Another important aspect is the adaptability of models across different scenarios. The IceBench tests demonstrate that models designed with adaptability in mind exhibit stronger performance across a diverse set of test conditions. This highlights the importance of versatility in real-world applications, where models must generalize effectively to unseen data. The ability to maintain robust performance under varying conditions is a key factor in ensuring the practical applicability of deep learning-based sea ice type classification. Additionally, our framework explores the influence of seasonal and geographic variability on model robustness. Models trained and tested on data from different seasons showed varying levels of performance, with those trained in transitional seasons like spring and summer demonstrating better generalization across all seasons. Similarly, geographical diversity in training data improved model performance, reinforcing the importance of incorporating datasets from multiple locations to enhance generalization capabilities.

Furthermore, IceBench also reveals the impact of architectural choices, particularly between pixel-based and patch-based models. While deeper networks improve accuracy at higher computational costs, patch-based models capture spatial dependencies more effectively, whereas pixel-based models excel in fine-grained classification. This trade-off between complexity, spatial resolution, and efficiency emphasizes the need for model selection based on task-specific and computational constraints. 
Beyond performance insights, IceBench provides answers to key research questions in sea ice classification, including the impact of downscaling resolution, patch size, data size, and data preparation methods on model performance. Additionally, feature importance analysis helps identify the most influential input features, guiding model interpretability and optimization. Furthermore, IceBench promotes standardization in model evaluation, enabling more meaningful comparisons and accelerating innovation in the field. Its findings also contribute to scalable and sustainable models, helping design models that balance effectiveness with computational efficiency to meet the growing demands of modern AI.


\vspace{6pt} 

\funding{This research was funded by the National Science Foundation (NSF) under grant numbers 2026962 and 2026865 }

\acknowledgments{The authors would like to acknowledge the support of the U.S. National Science Foundation under Grants No. 2026962 and 2026865, and the University of Colorado Denver's Alderaan cluster for providing computational resources.}


\isPreprints{
  \begin{adjustwidth}{-\extralength}{0cm}
  \printendnotes[custom] 
  \end{adjustwidth}
}{
  \printendnotes[custom]
}

\reftitle{References}

\isPreprints{}{
} 
\end{document}